\documentclass{article}

\usepackage{microtype}
\usepackage[dvips]{graphicx}
\graphicspath{ {./images/} }
\usepackage{booktabs} 

\usepackage{hyperref}

\newcommand{\Cov}{\mathrm{Cov}}

\usepackage[accepted]{icml2024}

\usepackage{amsmath}
\usepackage{amssymb}
\usepackage{mathtools}
\usepackage{amsthm}
\usepackage{xcolor}

\DeclareMathOperator{\Tr}{Tr}
\usepackage[capitalize,noabbrev]{cleveref}

\usepackage[font=footnotesize]{caption}
\usepackage{enumitem}
\usepackage{float}
\usepackage{stfloats}
\usepackage{nicefrac}
\usepackage{pdflscape}
\usepackage{subcaption}
\usepackage{tabularx}
\usepackage{url}
\usepackage[group-separator={,}]{siunitx}
\setlist[itemize]{noitemsep, topsep=0pt, parsep=0pt, partopsep=0pt}

\definecolor{turquoise}{RGB}{49,126,127}
\hypersetup{
    colorlinks=true,
    linkcolor=orange,
    urlcolor=turquoise,
    citecolor=orange
}
\theoremstyle{plain}
\newtheorem{theorem}{Theorem}[section]
\newtheorem{proposition}[theorem]{Proposition}

\theoremstyle{definition}

\theoremstyle{remark}

\usepackage[textsize=tiny]{todonotes}

\newcommand{\ourmethod}{Tripod}

\newcommand{\source}{s}
\newcommand{\data}{x}
\newcommand{\latent}{z}

\newcommand{\numof}[1]{{n_{#1}}}    
\newcommand{\spaceof}[1]{\mathcal{\MakeUppercase{#1}}}  
\newcommand{\var}[1]{{#1}}   

\newcommand{\genfun}{g}
\newcommand{\approxinv}[1]{\hat{{#1}}^{-1}}
\newcommand{\encoder}{\approxinv{\genfun}}
\newcommand{\decoder}{\hat{\genfun}}
\newcommand{\loss}[1]{\losssymbol_\text{#1}}
\newcommand{\losssymbol}{\mathcal{L}}
\newcommand{\dataset}{\mathcal{D}}
\newcommand{\weight}[1]{\weightsymbol_\text{#1}}
\newcommand{\weightsymbol}{\lambda}
\newcommand{\E}{\mathbb{E}}

\DeclareMathOperator*{\argmin}{arg\,min}
\DeclareMathOperator*{\Var}{Var}
\DeclarePairedDelimiter{\norm}{\lVert}{\rVert}
\DeclarePairedDelimiterX{\infdivx}[2]{(}{)}{%
  #1\;\delimsize\|\;#2%
}

\newcommand{\KL}{D_\text{KL} \infdivx}
\DeclareMathOperator{\stopgrad}{StopGrad}

\definecolor{turquoise}{RGB}{49,126,127}
\hypersetup{
    colorlinks=true,
    linkcolor=orange,
    urlcolor=turquoise,
    citecolor=turquoise
}

\icmltitlerunning{Tripod: Three Complementary Inductive Biases for Disentangled Representation Learning}

\begin{document}
\definecolor{LQ}{rgb}{1.00,0.38, 0.31}
\definecolor{KLM}{rgb}{1.00,0.66,0.13}
\definecolor{NHP}{rgb}{0.11,0.75,0.78}

\twocolumn[
\icmltitle{Tripod: Three Complementary Inductive Biases \\ for Disentangled Representation Learning}



\icmlsetsymbol{equal}{*}

\begin{icmlauthorlist}
\icmlauthor{Kyle Hsu}{equal,stanford}
\icmlauthor{Jubayer Ibn Hamid}{equal,stanford}
\icmlauthor{Kaylee Burns}{stanford}
\icmlauthor{Chelsea Finn}{stanford}
\icmlauthor{Jiajun Wu}{stanford}
\end{icmlauthorlist}

\icmlaffiliation{stanford}{Stanford University}
\icmlcorrespondingauthor{Kyle Hsu}{kylehsu@cs.stanford.edu}

\icmlkeywords{Machine Learning, ICML}

\vskip 0.3in
]



\printAffiliationsAndNotice{\icmlEqualContribution} 

\begin{abstract}
Inductive biases are crucial in disentangled representation learning for narrowing down an underspecified solution set. In this work, we consider endowing a neural network autoencoder with three select inductive biases from the literature: data compression into a grid-like latent space via quantization, collective independence amongst latents, and minimal functional influence of any latent on how other latents determine data generation. In principle, these inductive biases are deeply complementary: they most directly specify properties of the latent space, encoder, and decoder, respectively. In practice, however, naively combining existing techniques instantiating these inductive biases fails to yield significant benefits. To address this, we propose adaptations to the three techniques that simplify the learning problem, equip key regularization terms with stabilizing invariances, and quash degenerate incentives. The resulting model, Tripod, achieves state-of-the-art results on a suite of four image disentanglement benchmarks. We also verify that Tripod significantly improves upon its naive incarnation and that all three of its ``legs'' are necessary for best performance.

\end{abstract}

\section{Introduction} \label{sec:introduction}
How can we enable machine learning models to process raw perceptual signals into organized concepts similar to how humans do? This intuitively desirable goal has a well-studied formalization known as unsupervised disentangled representation learning: a model is tasked with teasing apart an unlabeled dataset's underlying sources (a.k.a. factors) of variation and representing them separately from one another, e.g., in independent components of a learned latent space. Beyond aesthetic motivations, achieving disentanglement is a potential stepping stone toward the holy grails of compositional generalization~\citep{bengio2013deep,wang2023measuring} and interpretability~\citep{rudin2022interpretable,zheng2022disentangled}. Despite this problem's importance, there persists a gulf between how well machine learning models and humans disentangle even on carefully curated datasets~\citep{gondal2019transfer,nie2019high}.

\begin{figure}[t]
    \centering
    \vspace{0pt}
    \includegraphics[width=0.45 \textwidth]{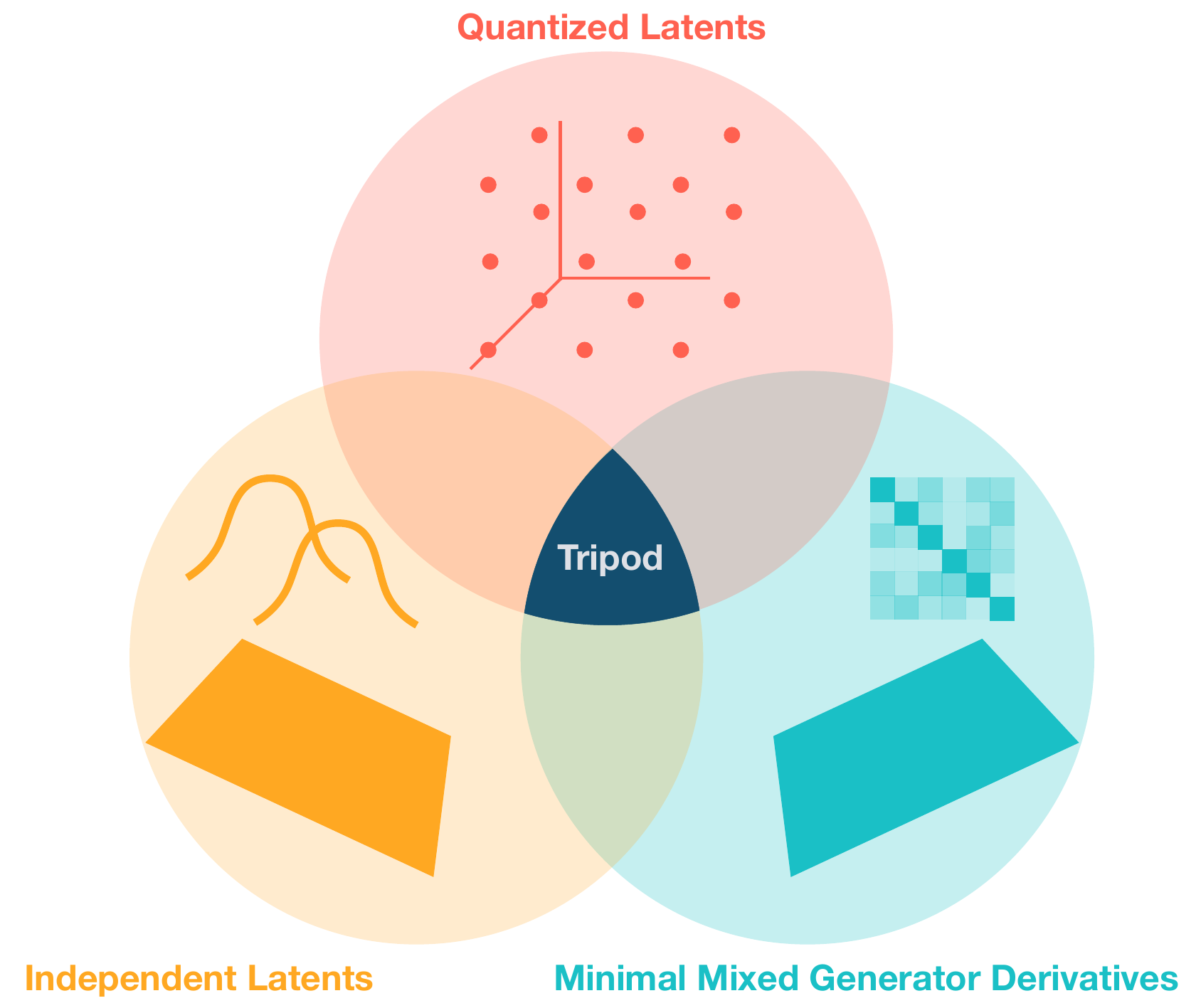}
    \caption{Each of the three inductive biases for disentanglement we consider in this work specifies a different set of preferred models (circles). In principle, using them in conjunction should more precisely specify the desired solution set and better recover models akin to the true data-generating process. Our method, \ourmethod{}, makes crucial modifications to these three ``legs'' to realize this synergy in practice. Code is available at \url{https://github.com/kylehkhsu/tripod}.}
    \label{fig:teaser}
    \vspace{-15pt}
\end{figure}

Inductive biases play a paramount role in enabling disentanglement: they help identify desired solutions amongst the space of all models that explain the data. In this work, we consider three select inductive biases proposed and validated in prior work to have some disentangling effect:
\begin{itemize}
    \item data compression into a grid-like latent space via quantization~\citep{hsu2023disentanglement}
    \item collective independence amongst latents~\citep{chen2018isolating,kim2018disentangling}
    \item minimal functional influence of any latent on how other latents determine data generation~\citep{peebles2020hessian}
\end{itemize}
While each of these desiderata has been shown to improve disentanglement, they achieve unsatisfactory performance in isolation.
Since establishing realistic sufficient conditions for identifiability has been a long-standing problem in disentanglement~\citep{locatello2019challenging,khemakhem2020variational,horan2021when}, it seems prudent to investigate the use of multiple inductive biases in conjunction to more precisely specify the desired solution set. 

The key insight this work offers is that the three aforementioned inductive biases, when integrated in a neural network autoencoding framework, are deeply complementary: they most directly specify properties of the latent space, encoder, and decoder, respectively. To elaborate, quantization of the latent space architecturally limits its channel capacity, necessitating efficient communication between the encoder and decoder. Meanwhile, the encoder shapes the joint density of the latents through how it ``places'' each datapoint, which must be done carefully to achieve collective independence. Finally, the decoder is responsible for minimizing the extent to which latents interact during data generation. Thus, while all three inductive biases ultimately influence the whole model, the mechanism by which each does so is distinct.

Unfortunately, naively combining existing instantiations of these inductive biases results in a model that performs poorly. We conjecture that one key cause of this is an increased difficulty in optimization, a well-known failure mode when juggling multiple objectives in deep learning. Our main technical contribution is a set of adaptations that ameliorate optimization difficulties by simplifying the learning problem, equipping key regularization terms with stabilizing invariances, and quashing degenerate incentives. We now briefly summarize these changes.

\textbf{Finite scalar latent quantization.} We leverage latent quantization to enforce data compression and encourage organization~\citep{hsu2023disentanglement}, but implement this via finite scalar quantization~\citep{mentzer2024finite} instead of dictionary learning~\citep{oord2017neural}. This fixes the codebook values and obviates two codebook learning terms in the objective, greatly stabilizing training early on and facilitating the optimization of the other inductive biases' regularization terms.

\textbf{Kernel-based latent multiinformation.} We adapt latent multiinformation regularization, originally proposed for variational autoencoders~\citep{chen2018isolating,kim2018disentangling}, to be compatible with deterministic encoders without needing an auxiliary discriminator.
We achieve this by a novel framing based on kernel density estimation.
This allows us to leverage well-known multivariate kernel design heuristics, such as incorporating each dimension's empirical standard deviation~\citep{silverman2018density}, in order to obtain density estimates that are more useful for multiinformation regularization. 

\textbf{Normalized Hessian penalty.} We derive a normalized version of the Hessian (off-diagonal) penalty~\citep{peebles2020hessian}. Unlike the original Hessian penalty, our regularization is invariant to dimensionwise rescaling of the decoder input (latent) and output (activation) spaces. This removes a key barrier to the fruitful application of data-generating mixed derivative regularization to autoencoder architectures, which we demonstrate for the first time; the original Hessian penalty was proposed for generative adversarial networks (GANs), in which the latent space is fixed.

The resulting method, \ourmethod{}, establishes a new state-of-the-art on a representative suite of four image disentanglement benchmarks~\citep{burgess20183d,gondal2019transfer,nie2019high} with an InfoMEC $=$ (InfoModularity, InfoCompactness, InfoExplicitness)~\citep{hsu2023disentanglement} of $(0.78, 0.59, 0.90)$ and a DCI $=$ (Disentanglement, Completeness, Informativeness)~\citep{eastwood2018framework} of $(0.64, 0.57, 0.93)$ in aggregate.
\ourmethod{} handidly outperforms methods from previous works that use just one of its component ``legs'' as well as ablated versions of itself that use two out of three ``legs'', validating the premise of using multiple inductive biases in conjunction. We also verify that \ourmethod{} disentangles much better than a naive incarnation combining previous instantiations of the three component inductive biases, thereby justifying our technical contributions.

\section{Preliminaries} \label{sec:preliminaries}
We begin by giving an overview of the specific disentangled representation learning problem we consider in this work.
Then, to contextualize our technical contributions and design decisions, we provide detailed descriptions of the previous methods we build upon.

\subsection{Disentangled Representation Learning} \label{sec:prelims_drl}
We consider the following disentangled representation learning problem statement inspired by nonlinear independent components analysis~\citep{hyv_arinen1999nonlinear,zheng2022identifiability,hsu2023disentanglement}. Given a dataset of paired samples of sources and data $\{(\source, \data)\}$ from a true data-generating process 
\begin{equation} \label{eq:nonlinear_ica}
    p(\var{\source}) = \prod_{i=1}^{\numof{\source}} p(\var{\source}_i), \enspace \var{\data} = \genfun(\var{\source}),
\end{equation}
where $\numof{\source}$ is the number of sources, our aim is to learn an encoder $\encoder : \spaceof{\data} \to \spaceof{\latent}$ and decoder $\decoder: \spaceof{\latent} \to \spaceof{\data}$ solely using the unlabelled data $\dataset = \{\data\}$ such that the latents $\latent$ recover the sources, thereby disentangling the data. To quantify disentanglement, we will use the InfoMEC~\citep{hsu2023disentanglement} and DCI~\citep{eastwood2018framework} metrics as estimated from samples $\{(\source, \latent = \encoder \circ \genfun(\source))\}$ from the joint source-latent distribution $p(\source, \latent)$. Both sets of metrics measure the modularity, compactness, and explicitness of the latents with respect to the sources. InfoM and D measure the extent to which each latent only contains information about one source (i.e., the extent to which the source-latent mapping is one-to-many); InfoC and C the extent to which each source is captured by only one latent (i.e., the extent to which the source-latent mapping is many-to-one); and InfoE and I the extent to which each source can be predicted from the latents with linear or random forest models, respectively. We will also qualitatively inspect models by visualizing the effect of intervening on latents prior to decoding.

\subsection{Inductive Biases from Prior Work}
\textbf{Latent quantization.} The true sources of variation are a neatly organized, highly compressed representation of the data. \citet{hsu2023disentanglement} propose to quantize continuous representations $\encoder(\data)$ onto a regular grid to mimic this structure and enforce compression. They accomplish this via a scalar form of vector quantization~(VQ; \citet{oord2017neural}):
\begin{equation} \label{eq:latent_quantization}
    \latent_j = \argmin_{e_{jl} \in \mathcal{E}_j} \; \lvert \encoder(\data)_j - e_{jl} \rvert, \enspace j=1,\dots,\numof{\latent},
\end{equation}
where $\{\mathcal{E}_j\}_{j=1}^{n_z}$ are the codebook values. These adapt via a ``quantization loss'' that amounts to dictionary learning, and the continuous values are simultaneously encouraged to collapse to their quantizations via a ``commitment loss'':
\begin{align} 
    \loss{quantize}(\{\mathcal{E}_j\}) &= \norm{\stopgrad(\encoder(\data)) - \latent}_2^2 \label{eq:vq_quantization}, \\
    \loss{commit}(\encoder) &= \norm{\encoder(\data) - \stopgrad(\latent)}_2^2 \label{eq:vq_commitment}.
\end{align}

\textbf{Latent multiinformation regularization.} Since the true sources are collectively independent~\eqref{eq:nonlinear_ica}, biasing the latents towards exhibiting this property should help with recovering something similar to the true generative process. One granular measure of independence is the multiinformation~\citep{studeny1998multiinformation} (a.k.a.~total correlation) of the latents,
\begin{equation} \label{eq:latent_multiinformation}
    \KL{q(\var{\latent})}{\textstyle \prod_{j=1}^\numof{\latent} q(\var{\latent}_j)},
\end{equation}
which vanishes with perfect collective independence. In a variational autoencoder~(VAE; \citet{kingma2014auto}), one can define $q(\latent) = \frac{1}{|\dataset|} \sum_{\data \in \dataset} q(\latent | \data)$ to be an ``aggregate posterior''.
Since naive Monte Carlo estimation of the multiinformation would require the entire dataset,
\citet{chen2018isolating} design a minibatch-weighted estimator for the expected log density:
\begin{equation} \label{eq:tc_estimator}
    \E_{\latent \sim q(\latent)} [ \log q(\latent) ] \approx \frac{1}{\numof{b}} \sum_{i_1=1}^{\numof{b}}  \log \frac{1}{\numof{b} |\dataset|} \sum_{i_2=1}^{\numof{b}} q\left(z^{(i_2)} | x^{(i_1)}\right) ,
\end{equation}
where $\numof{b}$ is the batch size and $|\dataset|$ is the dataset size. They handle the marginals $q(\latent_j)$ analogously, and add the estimated multiinformation as a regularization term to the VAE evidence lower bound objective. 

\textbf{Data-generating mixed derivative regularization.} \citet{peebles2020hessian} propose that, in a data-generating process that transforms latents into data, each latent should minimally affect how any other latent functionally influences the data. To accomplish this, they regularize the mixed derivatives of the generator in a generative adversarial network~\citep{goodfellow2014generative} (GAN):
\begin{equation}
    \min_{\decoder^{[k]}} \sum_{j_1 \neq j_2} \left(H^{[k]}_{j_1 j_2}\right)^2  
\end{equation}
where $\decoder^{[k]}$ denotes some generator activation or output dimension and $H^{[k]}$ is the corresponding Hessian with respect to the latents.
Naively implementing this via automatic differentiation is too computationally expensive, as it would involve taking third-order derivatives for first-order optimization schemes. Instead, \citet{peebles2020hessian} apply a Hutchinson-style unbiased estimator for the sum of the squared off-diagonal elements of a matrix~\citep{hutchinson1989stochastic}:
\begin{equation} \label{eq:hutchinson_off_diagonal}
    \Var_{v \sim \operatorname{Rademacher}(1)} [ v^\top H^{[k]} v ] = 2 \sum_{j_1 \neq j_2} 
    \left(H^{[k]}_{j_1j_2}\right)^2,
\end{equation}
as well as a central finite difference approximation for the second-order directional derivative:
\begin{equation} \label{eq:finite_differences}
    v^\top H^{[k]} v \approx \frac{\decoder^{[k]}(\latent + \epsilon v) - 2 \decoder^{[k]}(\latent) + \decoder^{[k]}(\latent - \epsilon v)}{\epsilon^2}.
\end{equation}

\section{The Three Legs of \ourmethod{}} \label{sec:methods}

In this section, we describe the design decisions we make in order to successfully meld latent quantization, latent multiinformation regularization, and data-generating mixed derivative regularization together in a single model. Our overarching design principle is as follows: we want each inductive bias to do its job with the lightest touch possible in order to avoid ``interference'' and ease optimization.

\subsection{Finite Scalar Latent Quantization (FSLQ)}
\label{sec:fslq}
\begin{figure}[t]
   \centering
   \includegraphics[width=0.48 \textwidth]{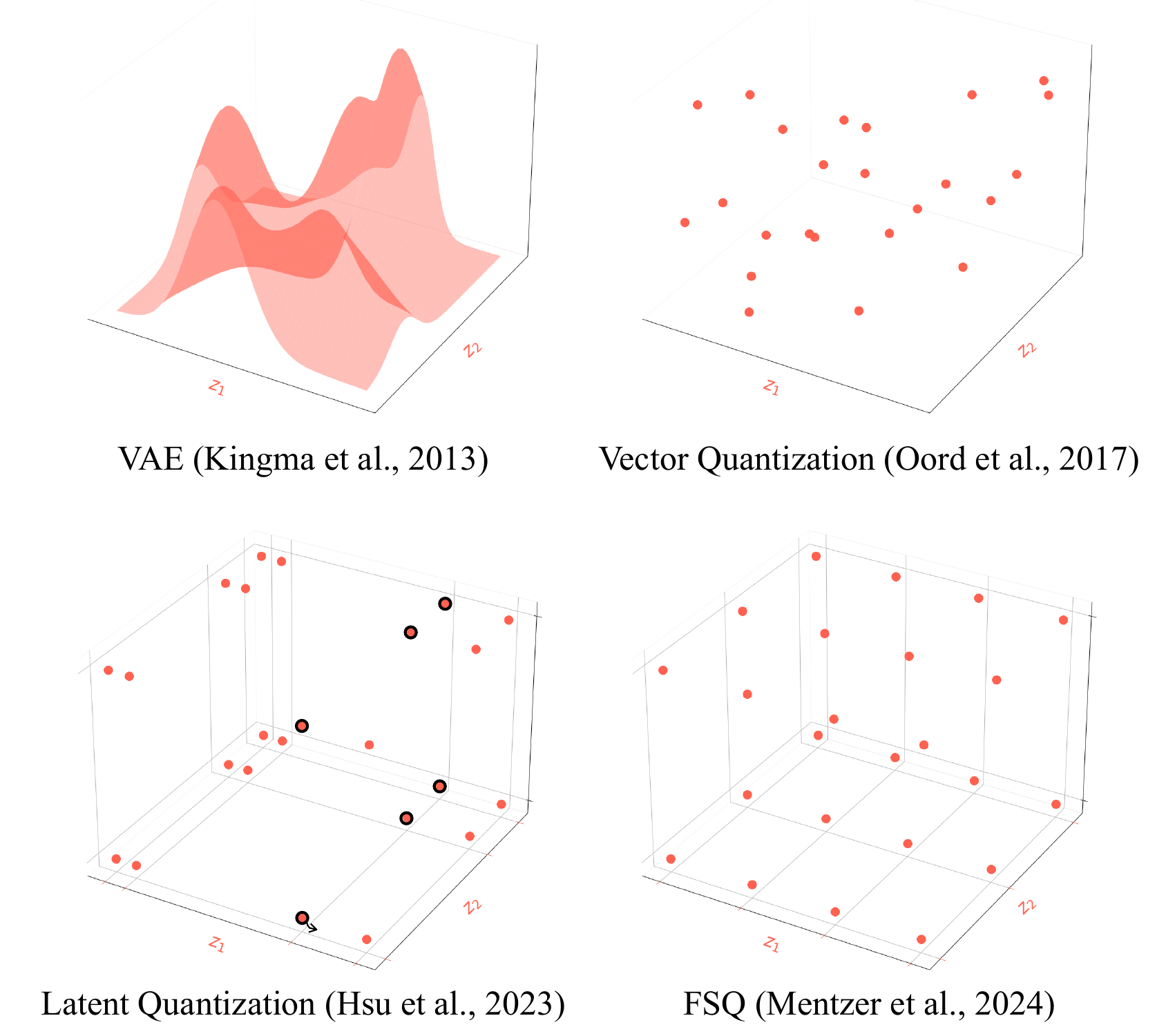}
   \vspace{-10pt}
   \caption{The evolution of discrete latent space structure in autoencoders. We use finite scalar quantization (bottom right) instead of latent quantization (bottom left) so that the codebook values need not be learned.}
   \label{fig:quantization}
   \vspace{-10pt}
\end{figure}
\begin{figure*}[t]
  \centering
  \includegraphics[width=.8\textwidth]{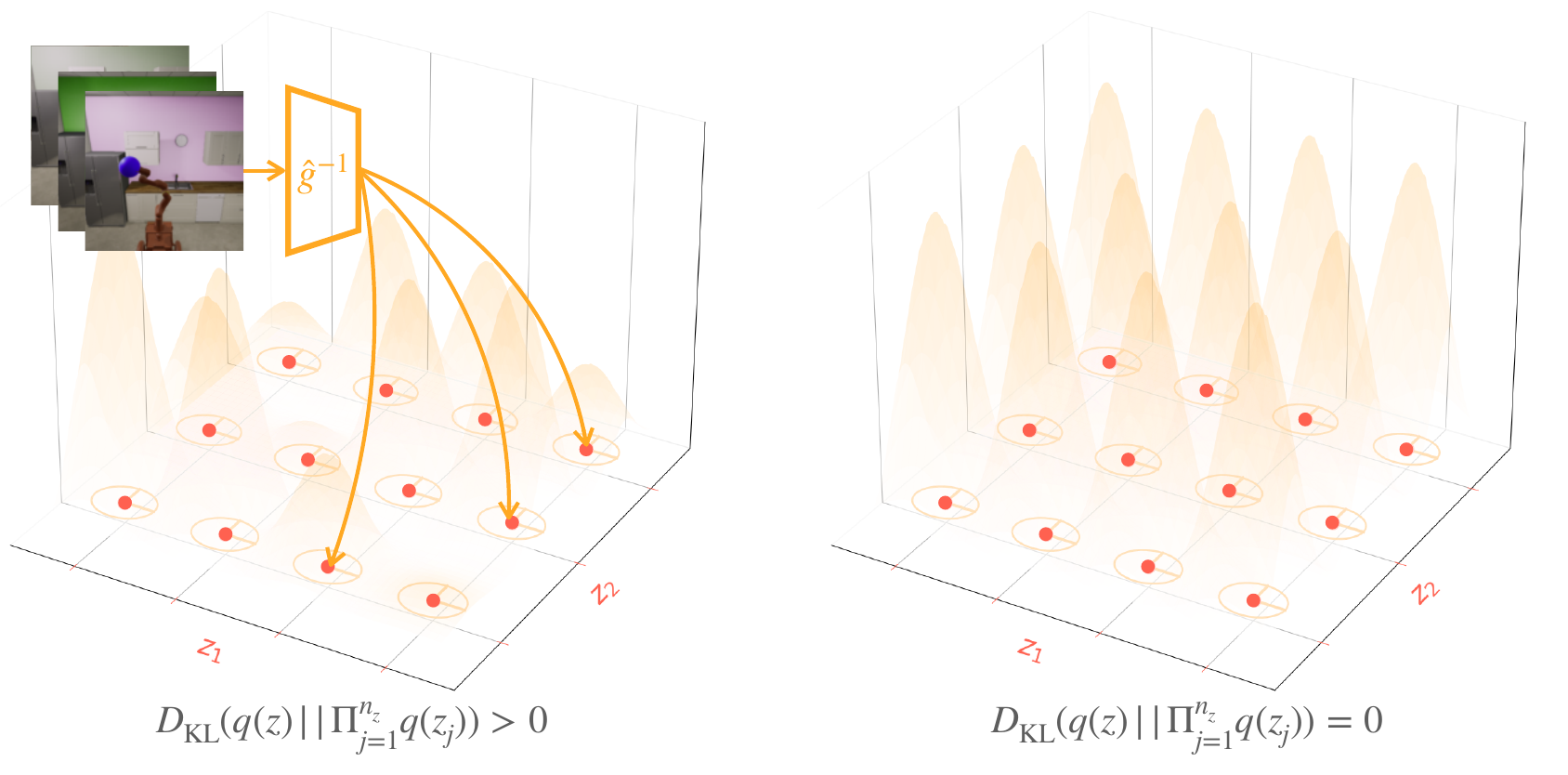}
  \vspace{-10pt}
  \caption{Kernel density estimation facilitates regularizing deterministic quantized latents from having nonzero multiinformation (left) towards collective independence (right). The multiinformation estimation smoothly depends on the latents through distances between samples~(\ref{eq:kde_joint}, \ref{eq:kde_marginal}). The smoothing matrix $S$~\eqref{eq:silvermans} is visualized with a level set (ellipse) of each latent sample's kernel density, and incorporates each dimension's scale (ellipse major and minor axes). The visualized joint densities illustrate the result of accumulating latent sample kernel densities at each grid point.}
  \label{fig:klm}
  \vspace{-10pt}
\end{figure*}

The scalar form of VQ that \citet{hsu2023disentanglement} use to instantiate latent quantization (LQ) requires a quantization loss~\eqref{eq:vq_quantization} for learning the discrete codebook values and a commitment loss~\eqref{eq:vq_commitment} to regularize the continuous values. We are looking to include other regularizers on the latents that more directly enable disentanglement, and these would play a bigger role in the objective if the VQ losses could be removed. The recently proposed finite scalar quantization ~(FSQ;~\citet{mentzer2024finite}) scheme identifies a recipe for doing this using fixed codebook values and judiciously chosen mappings pre- and post-quantization. We graphically depict VQ, LQ, and FSQ in \cref{fig:quantization}.

We implement a variant of FSQ that has the continuous and quantized latent spaces share the same ambient ``bounding box''. The continuous latent space is specified as $[-1,1]^\numof{\latent}$ by applying the hyperbolic tangent function to the output of the encoder network. Latent vectors are then linearly rescaled to $[0,\numof{q} - 1]^\numof{\latent}$, rounded elementwise to the nearest integer, and unscaled. Concretely, the quantization operation is
\begin{equation} \label{eq:fsq}
    \latent = \frac{2}{\numof{q} - 1} \operatorname{round}\left(
        \frac{\numof{q} - 1}{2} \left(\tanh\left(\encoder\left(\data\right)\right) + 1 \right)
    \right) - 1,
\end{equation}
where $\numof{q}$ is the number of discrete values in each dimension. The straight-through gradient trick~\citep{bengio2013estimating} is used to copy gradients across the nondifferentiable rounding operation.

FSQ directly obviates the quantization loss~\eqref{eq:vq_quantization} as the latent codebook is fixed to $\{-1, -1 + \frac{2}{\numof{q} - 1}, \dots, 1\}^\numof{\latent}$. \citet{mentzer2024finite} empirically show that the commitment loss~\eqref{eq:vq_commitment} also becomes unnecessary. These simplifications greatly stabilize the early periods of training and ``make room'' for the other two inductive biases, which we discuss next.

\subsection{Kernel-Based Latent Multiinformation (KLM)}

At face value, the idea of regularizing latent multiinformation~\eqref{eq:latent_multiinformation} \`{a} la \citet{chen2018isolating} appears to be incompatible with deterministic latents: since each ``posterior'' is a Dirac delta function, we have $q(\latent) = \frac{1}{|\dataset|} \sum_{\data \in \dataset} \delta(z - \encoder(x))$, which has a support of measure zero. We instead adopt a perspective of doing kernel density estimation (KDE) using a finite data sample: we specify a kernel-based smoothing of the Dirac deltas for the express purpose of obtaining smoothly parameterized density estimates that are amenable to gradient-based optimization. See \cref{fig:klm} for an illustration of our method. We make the standard choice of Gaussian kernel functions; concretely, we estimate the joint density as
\begin{equation} \label{eq:kde_joint}
    q(z) = \frac{1}{n_b} \sum_{i=1}^{n_b} \frac{1}{(2 \pi)^{\frac{\numof{\latent}}{2}} |S|^{\frac{1}{2}}}
     \exp\left( -\frac{1}{2} f(\latent - \latent^{(i)} ; S) \right),
\end{equation}
where $f(\latent' ; S) = \latent'^\top S^{-1} \latent'$ and $S$ is a smoothing matrix. In KDE, it is a common heuristic to incorporate the empirical standard deviation of each dimension $\sigma_j$ into the kernel smoothing parameters~\citep{silverman2018density}. 
This specifies an invariance to the dimensionwise scaling, facilitating, e.g., latent shrinkage without adversely affecting multiinformation estimation. Specifically, to estimate the joint density $q(\latent)$, we use Silverman's rule of thumb for each nonzero element of the diagonal smoothing matrix: 
\begin{equation} \label{eq:silvermans}
    S_{jj} = \left(\frac{4}{(\numof{\latent} + 2)\numof{b}}\right)^\frac{2}{\numof{\latent} + 4} \sigma_j^2.
\end{equation}
For estimating the marginal densities, we again use $\sigma_j$ as the smoothing parameter:
\begin{equation} \label{eq:kde_marginal}
    q(z_j) = \frac{1}{n_b}\sum_{i=1}^{n_b} \frac{1}{\sqrt{2\pi} \sigma_j}\exp\left( - \frac{1}{2}\left( \frac{z_j - z_j^{(i)}}{\sigma_j}\right)^2\right).
\end{equation}
We now have satisfactory approximations for the densities involved in computing latent multiinformation~\eqref{eq:latent_multiinformation}.
Our batch log joint density estimator,
\begin{equation}
    \E_{\latent \sim q(\latent)} [ \log q(\latent) ] \approx \frac{1}{\numof{b}} \sum_{i_1=1}^{\numof{b}} \log \frac{1}{\numof{b}} \sum_{i_2=1}^{\numof{b}} K_S\left(z^{(i_2)} - z^{(i_1)}\right)
\end{equation}
ultimately takes a highly similar form to the minibatch-weighted estimator~\eqref{eq:tc_estimator} of \citet{chen2018isolating}, except the pairwise interaction between samples arises from kernel-based smoothing rather than uncertainty in posterior inference. We also avoid the inclusion of the dataset size due to not using an importance sampling derivation. Cosmetic differences aside, the two methods are deeply related: one can view the design of the variational posterior $q$ as analogous to the design of the kernel $K_S$. We remark that KLM is designed for deterministic latents taking on values in $\mathbb{R}^{\numof{\latent}}$. This includes our quantized latents as a specific case thanks to the design of the quantized latent space in \cref{sec:fslq}. 

As an alternative to kernel-based smoothing, we can consider treating the quantized latents as categorical variables, but we find that the required machinery, e.g., Gumbel-Softmax reparameterization~\citep{jang2017categorical,maddison2017concrete}, is significantly more unwieldy to use in practice compared to kernel density estimation.

\subsection{Normalized Hessian Penalty (NHP)}

\begin{algorithm*}[t]
   \caption{Pseudocode for the \ourmethod{} objective. We use $\numof{b}=64, \numof{p}=2, \epsilon=0.1$ throughout and tune $(\lambda_\text{KLM}, \lambda_\text{NHP})$.}
   \label{alg:tripod}
\begin{algorithmic}[1]
    \STATE {\bfseries given:} batch size $\numof{b}$, data $\{\data^{(i)}\}_{i=1}^{\numof{b}}$, encoder $\encoder$, decoder $\decoder$, number of perturbations $\numof{p}$, perturbation parameter $\epsilon$ \\
    {\color{white}\bfseries given:} regularization weights $(\lambda_\text{KLM}, \lambda_\text{NHP})$ 
    \FOR{$i \in [\numof{b}]$}
        \STATE $c^{(i)} \gets \encoder(\data^{(i)})$ \hfill
            \COMMENT {encode data into continuous latent vectors}
        \STATE {\color{black} $\latent^{(i)} \gets \operatorname{Quantize}\left( c^{(i)} \right) $}  \hfill
            \COMMENT {apply finite scalar quantization~\eqref{eq:fsq}}
    \ENDFOR
    \STATE {\color{LQ} $\loss{reconstruction} \gets \frac{1}{\numof{b}} \sum_{i=1}^\numof{b} \operatorname{BinaryCrossEntropy}\left(\decoder(\latent^{(i)}), \data^{(i)} \right)$}
    
    \STATE $\sigma_j \gets \operatorname{std}\left(c^{(1)}_j, \dots, c^{(\numof{b})}_j \right) \ \forall j \in [\numof{\latent}]$ \hfill
        \COMMENT {calculate the empirical standard deviation of each latent dimension}
    
    \STATE {\color{black} $S \gets \operatorname{Silverman's}\left(\sigma_1,\dots,\sigma_{\numof{\latent}}, \numof{b}, \numof{z}\right)$} \hfill
        \COMMENT {form joint density smoothing matrix~\eqref{eq:silvermans}}
    \FOR{$i \in [\numof{b}]$}
        \STATE {\color{black} $q^{(i)} \gets \operatorname{KDE}\left(\latent^{(i)}; \latent^{(1)}, \dots, \latent^{(\numof{b})}, S\right)$} \hfill
        \COMMENT {joint KDE~\eqref{eq:kde_joint}}
        \STATE {\color{black} $q_j^{(i)} \gets \operatorname{KDE}\left(\latent_j^{(i)}; \latent^{(1)}_j, \dots, \latent^{(\numof{b})}_j, \sigma_j \right) \forall j \in [\numof{\latent}]$} \hfill
        \COMMENT {marginal KDEs~\eqref{eq:kde_marginal}}
    \ENDFOR
    \STATE {\color{KLM} $\loss{latent multiinformation} \gets \frac{1}{\numof{b}} \sum_{i=1}^\numof{b} \left(\log q^{(i)} - \sum_{j=1}^\numof{\latent} \log q_j^{(i)} \right)$}

    \FOR{$i \in [\numof{b}], k \in \{\text{regularized decoder activation dimensions}\}$}
        \FOR{$l \in \numof{p}$}
            \STATE {\color{black} $v^{(i)}_{jkl} \gets \sigma_j \operatorname{SampleRademacher}(1) \ \forall j \in [\numof{\latent}] $} \hfill
                \COMMENT {scale-adjusted sampling~(\cref{prop:nhp_calculation})}
            \STATE {\color{black} $w^{(i)}_{jkl} \gets \sigma_j \operatorname{SampleNormal}(0, 1)  \ \forall j \in [\numof{\latent}] $}
            \STATE {\color{black} $\text{numer}f^{(i)}_{kl} \gets \operatorname{FiniteDifferences}\left(\decoder^{[k]}, \latent^{(i)}, \epsilon, v^{(i)}_{kl} \right)  $} \hfill
                \COMMENT {estimate curvature~\eqref{eq:finite_differences}}
            \STATE {\color{black} $\text{denom}f^{(i)}_{kl} \gets \operatorname{FiniteDifferences}\left(\decoder^{[k]}, \latent^{(i)}, \epsilon, w^{(i)}_{kl} \right) $}
        \ENDFOR
        \STATE {\color{black} $\text{numer}f^{(i)}_{k} \gets \operatorname{var}\left(\text{numer}f^{(i)}_{k1}, \dots, \text{numer}f^{(i)}_{k\numof{p}} \right)  $} \hfill
            \COMMENT {calculate empirical variance across perturbations~\eqref{eq:nhp_calculation}}
        \STATE {\color{black} $\text{denom}f^{(i)}_{k} \gets \operatorname{var}\left(\text{denom}f^{(i)}_{k1}, \dots, \text{denom}f^{(i)}_{k\numof{p}} \right)  $} \hfill
    \ENDFOR
   \STATE {\color{NHP} $\loss{normalized Hessian penalty} \gets \frac{1}{\numof{b}} \sum_{i=1}^\numof{b} \frac{\sum_k \text{numer}f^{(i)}_{k}}{\sum_k \text{denom}f^{(i)}_{k}}$}
    \STATE {\bfseries return:} 
        $\loss{} \gets $ 
        {\color{LQ} $\loss{reconstruction}$} $+$ {\color{KLM} $\lambda_\text{KLM} \loss{latent multiinformation}$} $+$ 
        {\color{NHP} $\lambda_\text{NHP} \loss{normalized Hessian penalty}$}

\end{algorithmic}
\end{algorithm*}

\begin{figure}[t]
  \centering
  \includegraphics[width=0.45\textwidth]{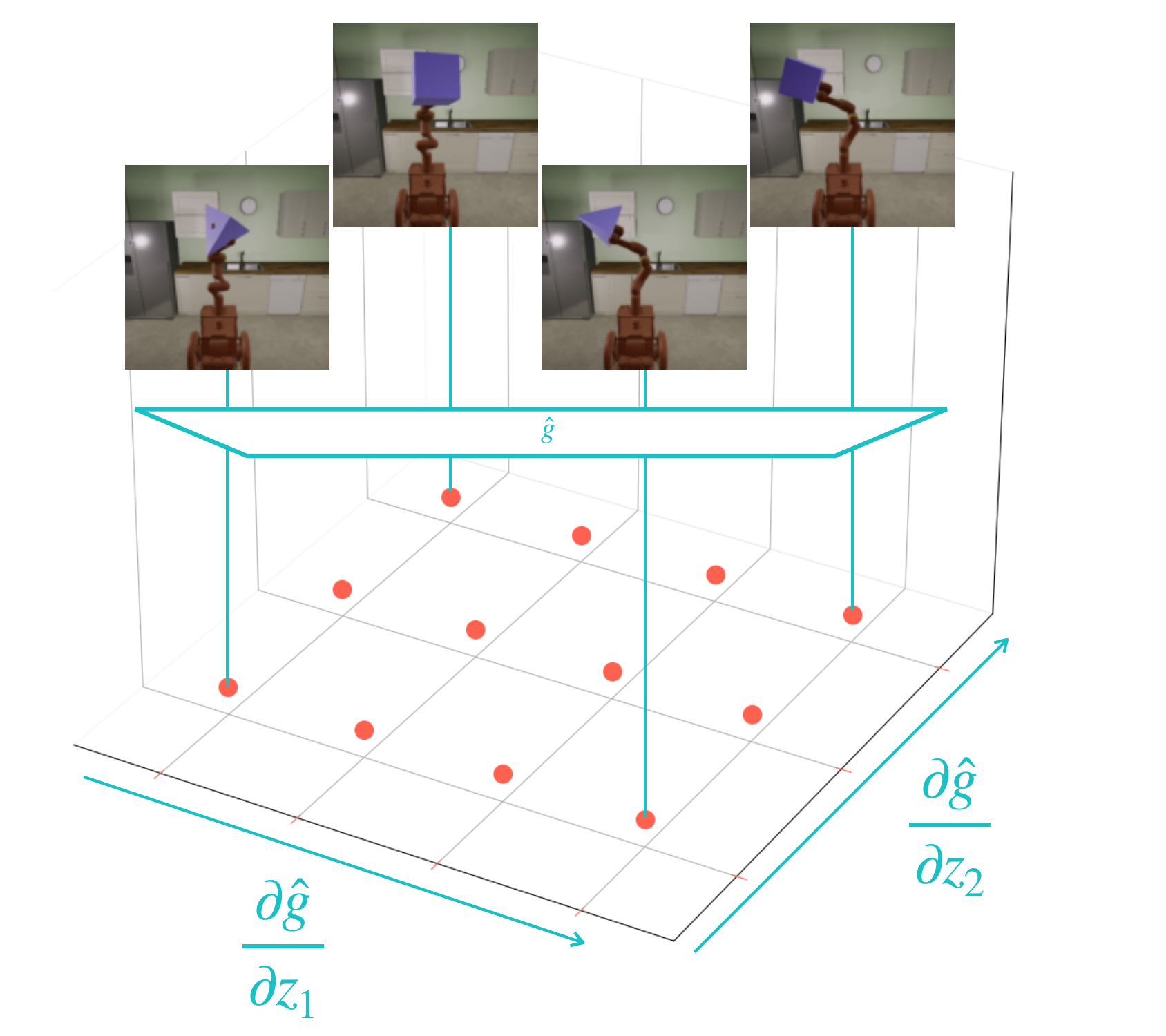}
  \caption{The Hessian penalty is supposed to specify a preference for decoders, such as the one depicted, in which change along one latent (object shape) minimally affects how another latent (horizontal end-effector position) influences data generation. We modify the Hessian penalty to quash degenerate solutions that compromise this intended outcome in autoencoders.}
  \vspace{-15pt}
  \label{fig:nhp}
\end{figure}

The original Hessian penalty \eqref{eq:hutchinson_off_diagonal} amounts to reducing the magnitudes of mixed derivatives in a learned data-generating function~(\cref{fig:nhp}). Unfortunately, this can be trivially achieved by scaling down activations or scaling up latents, circumventing the intended effect of making the Hessians more diagonal. Indeed, the former degeneracy is likely why \citet{peebles2020hessian} find it important to use normalization layers immediately preceding activations used for regularization: in experiments with the original Hessian penalty, we find that it causes the norms of regularized activations to decrease. More importantly, the latter degeneracy may be why the Hessian penalty has not seen fruitful application in autoencoders: the decoder input space is variable and hence susceptible, whereas the input space of a GAN's generator is fixed.

We would like to rule out trivial scaling-based solutions to data-generating mixed derivative regularization by making the regularization term invariant to the scale of any individual input (latent) or output (activation). We endow the Hessian penalty with these properties by incorporating the standard deviations of the latents in each second derivative and normalizing by an aggregation of all second derivatives. This is formalized in Proposition \ref{prop:hp_vs_nhp}.

\begin{proposition} \label{prop:hp_vs_nhp}
The Hessian penalty
    \begin{equation}
        \label{eq:vanilla_hessian_penalty}
        \sum_{j_1\neq j_2} \left(H^{[k]}_{j_1 j_2}\right)^2
    \end{equation}
can be reduced by scaling down $\decoder{}_k$ or scaling up any $z_j, j \in [\numof{\latent}]$, and vice versa. In contrast, the normalized Hessian penalty
    \begin{equation}
        \label{eq:normalized_hessian_penalty}
        \frac{\sum_{j_1\neq j_2} \left(H^{[k]}_{j_1 j_2} \sigma_{j_1} \sigma_{j_2}\right)^2}{\sum_{j_1,j_2} \left(H^{[k]}_{j_1 j_2} \sigma_{j_1} \sigma_{j_2}\right)^2}  
    \end{equation}
is invariant to the scaling of $\decoder^{[k]}$ and $\latent_j \ \forall j \in [\numof{\latent}]$.
\end{proposition}
\begin{proof}
    See \cref{app:hp_vs_nhp_proof}.
\end{proof}

However, incorporating the latent standard deviations into each term is nontrivial since the Hutchinson-style estimation~\eqref{eq:hutchinson_off_diagonal} never explicitly forms any of the terms in the sum. We also need a way of estimating the denominator of \eqref{eq:normalized_hessian_penalty}, which includes the Hessian's squared diagonal entries. Fortunately, each can be achieved with a judicious change to the sampling distribution, as we show in \cref{prop:nhp_calculation}.
\begin{proposition} \label{prop:nhp_calculation}
Let $v$ and $w$ be random vectors where $v_j \sim \operatorname{Rademacher}(\sigma_j)$ and $w_j \sim \mathcal{N}(0, \sigma_j^2)$. Then the normalized Hessian penalty can be computed as
    \begin{equation} \label{eq:nhp_calculation}
        \frac{\sum_{j_1 \neq j_2} \left(H^{[k]}_{j_1 j_2} \sigma_{j_1} \sigma_{j_2}\right)^2} { \sum_{j_1, j_2} \left(H^{[k]}_{j_1 j_2} \sigma_{j_1} \sigma_{j_2}\right)^2} = \frac{\Var\left[v^T H^{[k]} v\right]}{\Var\left[w^T H^{[k]} w\right]} .
    \end{equation} 
\end{proposition}
\begin{proof}
    See \cref{app:nhp_calculation_proof}.
\end{proof}
Using a central finite difference approximation~\eqref{eq:finite_differences} for the second-order directional derivatives~\eqref{eq:nhp_calculation}, we are now able to estimate the normalized Hessian penalty~\eqref{eq:normalized_hessian_penalty} just with forward passes through the decoder $\decoder(z)$. Compared to the original Hessian penalty, this incurs twice as many forward passes per optimization step.

\subsection{Implementation Details}
In \cref{alg:tripod}, we provide pseudocode for computing the Tripod objective. There are a few implementation details worth noting. We compute the latents' empirical standard deviation based on the \emph{continuous} values to avoid obtaining a value of zero for a batch. Substituting the statistics of one for the other is facilitated by tying together the continuous and quantized latent spaces as outlined in \cref{sec:fslq}. However, for the kernel density estimates and finite differences, we use the quantized latents. Also, due to the low number of perturbations (2) used for the curvature approximations, we find it more stable to separately aggregate the numerators and denominators of the normalized Hessian penalties for each decoder activation dimension before division. For the degree of quantization $\numof{q}$, we use a fixed value of $12$, except when we ablate this hyperparameter specifically. Finally, while we explicitly write out nested loops in \cref{alg:tripod} to maximize clarity, in code these are vectorized for the sake of efficiency.

{
\setlength{\tabcolsep}{1.6pt}
\begin{table*}[t]
    \caption{\ourmethod{} achieves state-of-the-art disentanglement as quantified by InfoMEC and DCI (the former is re-ordered to align with the latter). See \cref{sec:experiments_quantitative} for detailed commentary.}
    \vspace{-7pt}
    \label{tab:main}
    \centering
    \small
    \begin{tabular}{lccccccccccccccc}
        \toprule
        model & \multicolumn{3}{c}{aggregated} & \multicolumn{3}{c}{Shapes3D} & \multicolumn{3}{c}{MPI3D} & \multicolumn{3}{c}{Falcor3D} & \multicolumn{3}{c}{Isaac3D} \\
        \midrule
        & \multicolumn{15}{c}{$\text{InfoMEC} := (\text{InfoM} \ \text{InfoC} \ \text{InfoE})$} \\
        \cmidrule{2-16}
        $\beta$-TCVAE & $(0.62$ & $\mathbf{0.57}$ & $0.77)$ & $(0.68$ & $0.55$ & $\mathbf{0.98})$ & $(0.45$ & $0.42$ & $0.61)$ & $(0.71$ & $\mathbf{0.71}$ & $0.72)$ & $(0.65$ & $0.61$ & $0.78)$ \\
QLAE & $(0.68$ & $0.43$ & $0.88)$ & $(0.86$ & $0.45$ & $\mathbf{1.00})$ & $(0.52$ & $0.46$ & $0.75)$ & $(0.62$ & $0.39$ & $\mathbf{0.82})$ & $(0.72$ & $0.42$ & $\mathbf{0.94})$ \\
Tripod (naive) & $(0.68$ & $0.44$ & $\mathbf{0.90})$ & $(0.83$ & $0.45$ & $\mathbf{1.00})$ & $(0.61$ & $0.47$ & $0.84)$ & $(0.64$ & $0.38$ & $0.81)$ & $(0.65$ & $0.46$ & $0.93)$ \\
Tripod (ours) & $(\mathbf{0.78}$ & $\mathbf{0.59}$ & $\mathbf{0.90})$ & $(\mathbf{0.94}$ & $\mathbf{0.59}$ & $\mathbf{1.00})$ & $(\mathbf{0.64}$ & $\mathbf{0.53}$ & $0.84)$ & $(\mathbf{0.72}$ & $0.56$ & $\mathbf{0.82})$ & $(\mathbf{0.84}$ & $\mathbf{0.68}$ & $\mathbf{0.95})$ \\
Tripod w/o NHP & $(0.70$ & $0.48$ & $0.89)$ & $(0.85$ & $0.46$ & $\mathbf{1.00})$ & $(0.60$ & $0.50$ & $0.81)$ & $(0.59$ & $0.40$ & $0.81)$ & $(0.75$ & $0.57$ & $0.93)$ \\
Tripod w/o KLM & $(0.73$ & $0.50$ & $\mathbf{0.90})$ & $(0.89$ & $\mathbf{0.57}$ & $\mathbf{1.00})$ & $(0.57$ & $0.50$ & $0.80)$ & $(\mathbf{0.74}$ & $0.54$ & $\mathbf{0.82})$ & $(0.72$ & $0.38$ & $\mathbf{0.96})$ \\
Tripod w/ finer quantization & $(0.56$ & $0.46$ & $\mathbf{0.92})$ & $(0.69$ & $0.48$ & $\mathbf{1.00})$ & $(0.43$ & $0.40$ & $\mathbf{0.97})$ & $(0.54$ & $0.41$ & $\mathbf{0.84})$ & $(0.57$ & $0.54$ & $0.87)$ \\

        \midrule
        & \multicolumn{15}{c}{$\text{DCI} := (\text{D} \ \text{C} \ \text{I})$} \\
        \cmidrule{2-16}
        $\beta$-TCVAE & $(0.44$ & $0.38$ & $0.89)$ & $(0.64$ & $0.51$ & $\mathbf{1.00})$ & $(0.29$ & $0.26$ & $0.80)$ & $(0.42$ & $0.37$ & $\mathbf{0.86})$ & $(0.39$ & $0.36$ & $0.89)$ \\
QLAE & $(0.55$ & $0.43$ & $\mathbf{0.92})$ & $(\mathbf{0.79}$ & $0.58$ & $\mathbf{1.00})$ & $(0.42$ & $0.37$ & $0.82)$ & $(0.40$ & $0.31$ & $\mathbf{0.88})$ & $(0.60$ & $0.46$ & $\mathbf{0.98})$ \\
Tripod (naive) & $(0.57$ & $0.45$ & $\mathbf{0.93})$ & $(0.74$ & $0.55$ & $\mathbf{1.00})$ & $(0.46$ & $0.40$ & $\mathbf{0.85})$ & $(0.45$ & $0.34$ & $\mathbf{0.87})$ & $(0.62$ & $0.50$ & $\mathbf{0.99})$ \\
Tripod (ours) & $(\mathbf{0.64}$ & $\mathbf{0.57}$ & $\mathbf{0.93})$ & $(\mathbf{0.80}$ & $\mathbf{0.65}$ & $\mathbf{1.00})$ & $(\mathbf{0.54}$ & $\mathbf{0.48}$ & $\mathbf{0.86})$ & $(0.49$ & $0.47$ & $\mathbf{0.88})$ & $(\mathbf{0.72}$ & $\mathbf{0.67}$ & $\mathbf{0.99})$ \\
Tripod w/o NHP & $(0.56$ & $0.46$ & $\mathbf{0.92})$ & $(0.76$ & $0.56$ & $\mathbf{1.00})$ & $(0.48$ & $0.42$ & $\mathbf{0.86})$ & $(0.39$ & $0.30$ & $0.85)$ & $(0.63$ & $0.58$ & $\mathbf{0.98})$ \\
Tripod w/o KLM & $(0.60$ & $0.51$ & $\mathbf{0.92})$ & $(0.76$ & $0.62$ & $\mathbf{1.00})$ & $(0.49$ & $0.42$ & $0.83)$ & $(\mathbf{0.52}$ & $\mathbf{0.50}$ & $\mathbf{0.88})$ & $(0.62$ & $0.48$ & $\mathbf{0.99})$ \\
Tripod w/ finer quantization & $(0.51$ & $0.53$ & $0.88)$ & $(0.73$ & $\mathbf{0.64}$ & $\mathbf{1.00})$ & $(0.42$ & $0.40$ & $0.82)$ & $(0.40$ & $0.45$ & $0.83)$ & $(0.47$ & $0.63$ & $0.88)$ \\

        \bottomrule
  \end{tabular}
  \vspace{-5pt}
\end{table*}
}

\section{Experiments}
We design our experiments\footnote{Code is available at \url{https://github.com/kylehkhsu/tripod}.} to answer the following questions:
\begin{itemize}
    \item How well does \ourmethod{} disentangle compared to strong methods from prior work?
    \item Does \ourmethod{} significantly improve upon its naive incarnation?
    \item Which of the three ``legs'' of \ourmethod{} are important?
\end{itemize}

\subsection{Experimental Protocol}
We benchmark on four established image datasets with ground-truth source labels that facilitate quantitative evaluation: Shapes3D~\citep{burgess20183d}, MPI3D ~\citep{gondal2019transfer}, Falcor3D~\citep{nie2019high}, and Isaac3D~\citep{nie2019high}. Each dataset is constructed to satisfy the assumptions of the disentangled representation learning problem assumptions~\eqref{eq:nonlinear_ica}: sources are collectively independent, and data generation is near-noiseless. MPI3D is collected with a real-world robotics apparatus, whereas the other three are synthetically rendered. Further dataset details are presented in \cref{app:datasets}. We follow prior work in considering a statistical learning problem: we use the entire dataset for unsupervised training and evaluate on a subset of $10,000$ samples~\citep{locatello2019challenging}. We filter checkpoints for adequate reconstruction: we threshold based on the peak signal-to-noise ratio (PSNR) for each dataset at which reconstruction errors are imperceptible~(\cref{app:datasets}). We then compute the InfoMEC and DCI metrics introduced in \cref{sec:prelims_drl} and report for each run the results given by the checkpoint with the best InfoM.

For prior methods, we consider two works that introduced two of the inductive biases we use: $\beta$-total correlation variational autoencoding~($\beta$-TCVAE; \citet{chen2018isolating}) and quantized latent autoencoding~(QLAE; \citet{hsu2023disentanglement}). Since we use expressive convolutional neural network architectures for the encoder and decoder~\citep{dhariwal2021diffusion}, we implement these methods based on their reference open-source repositories in our own codebase for an apples-to-apples comparison. For the naive version of \ourmethod{}, we use VQ-style latent quantization as in QLAE, fixed smoothing parameters for kernel-based latent multiinformation regularization, and the vanilla Hessian penalty (with activation normalization). For all of the above methods, we tune key hyperparameters~(\cref{app:hyperparameters}) per dataset over $2$ seeds before switching to a different evaluation set and running $3$ more seeds.

\subsection{Quantitative Results} \label{sec:experiments_quantitative}
Main quantitative results are summarized in \cref{tab:main}.

\begin{figure*}
    {\scriptsize
    \begin{subfigure}{0.48\textwidth}
        \begin{tabularx}{0.97\textwidth}{*{18}{>{\centering\arraybackslash}X}}
            $z_0$ & \textcolor{red}{$z_1$} & $z_2$ & \textcolor{red}{$z_3$} & \textcolor{red}{$z_4$} & $z_5$ & $z_6$ & $z_7$ & $z_8$ & $z_9$ & $z_{10}$ & $z_{11}$ & $z_{12}$ & $z_{13}$ & $z_{14}$ & $z_{15}$ & $z_{16}$ & $z_{17}$ \\
        \end{tabularx}
    \end{subfigure}
    \hfill
    \begin{subfigure}{0.48\textwidth}
        \begin{tabularx}{0.97\textwidth}{*{18}{>{\centering\arraybackslash}X}}
            $z_0$ & $z_1$ & $z_2$ & $z_3$ & $z_4$ & $z_5$ & $z_6$ & $z_7$ & $z_8$ & $z_9$ & $z_{10}$ & $z_{11}$ & $z_{12}$ & $z_{13}$ & $z_{14}$ & $z_{15}$ & $z_{16}$ & $z_{17}$ \\
        \end{tabularx}
    \end{subfigure}
    }
    \begin{subfigure}{0.48\textwidth}
        \includegraphics[width=1.0\linewidth]{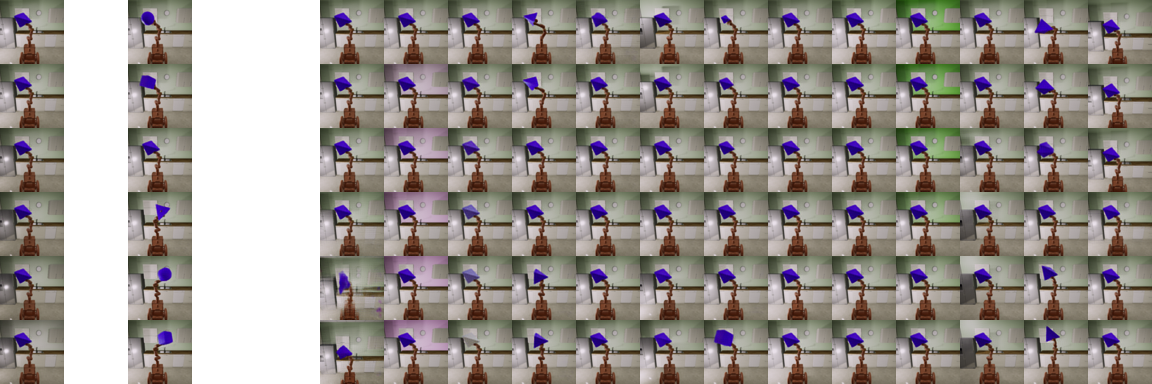}
    \end{subfigure}
    \hfill
    \begin{subfigure}{0.48\textwidth}
        \includegraphics[width=1.0\linewidth]{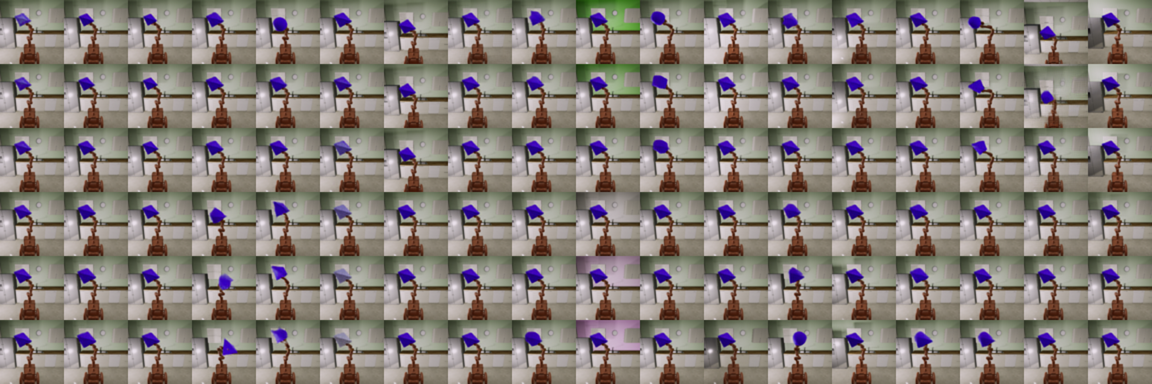}
    \end{subfigure}
    \begin{subfigure}{0.48\textwidth}
        \centering
        \includegraphics[width=1.0\linewidth]{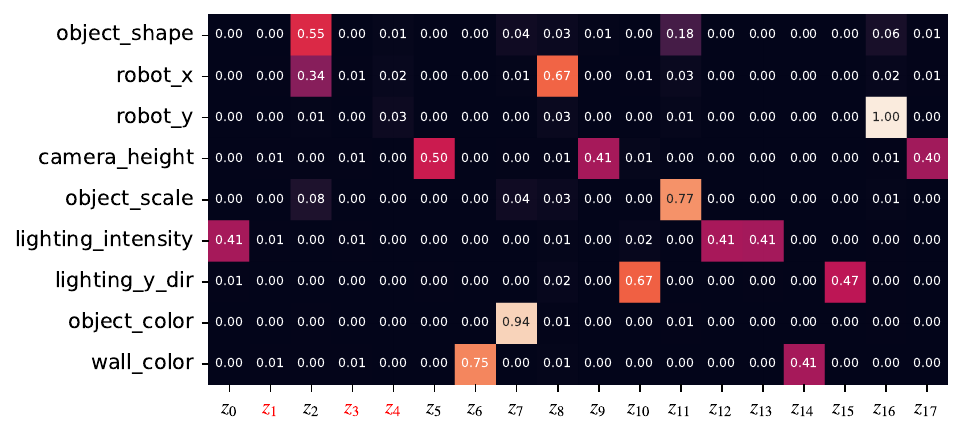}
        \caption*{\ourmethod{}}
    \end{subfigure}
    \hfill
    \begin{subfigure}{0.48\textwidth}
        \centering
        \includegraphics[width=1.0\linewidth]{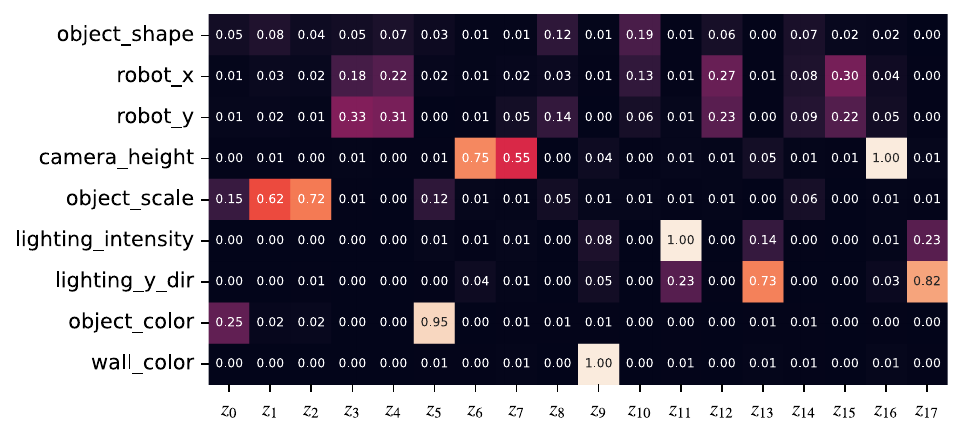}
        \caption*{Naive \ourmethod{}}
    \end{subfigure}
    \vspace{-5pt}
    \caption{Qualitative study of \ourmethod{} and naive \ourmethod{} on Isaac3D. Decoded latent interventions (top): in each column, we encode an image and visualize the effect of intervening on a single latent on decoding by varying its value in a linear interpolation in that latent's range. Normalized mutual information heatmaps (bottom): this acts as an ``answer key'' to what the observed qualitative changes in a column should be when considering the entire dataset. Red latents are inactive and corresponding columns are removed from the latent intervention visualizations. For more qualitative results, see \cref{app:qualitative_results}.}
    \label{fig:qualitative_isaac_main}
    \vspace{-10pt}
\end{figure*}

\textbf{Comparison of \ourmethod{} with $\beta$-TCVAE, QLAE, and naive \ourmethod{}.} \ourmethod{} outperforms $\beta$-TCVAE in both modularity metrics (InfoM and D), the DCI compactness metric C, and both explicitness metrics (InfoE and I). The total correlation regularization in $\beta$-TCVAE is analogous to the kernel-based latent multiinformation (KLM) regularization in \ourmethod{}; since this directly optimizes for compactness, it is no surprise that the two methods are competitive in InfoC. However, the substantial difference in InfoE indicates that the sources are more linearly predictable from \ourmethod{} latents than $\beta$-TCVAE latents.

In contrast, \ourmethod{} handidly outperforms QLAE in all modularity and compactness metrics while achieving parity in both explicitness metrics. This suggests that latent quantization, a property common to both methods, is important to achieve high explicitness. The difference in modularity and compactness is explained by the lack of the other two inductive biases, KLM and NHP, in QLAE.

\ourmethod{} dominates naive \ourmethod{} on every disentanglement metric for each dataset. This validates the utility of our technical contributions. Indeed, naive \ourmethod{} is virtually indistinguishable from QLAE in aggregate, demonstrating that our specific modifications to the three inductive biases are necessary to achieve a synergistic disentangling effect. We highlight that the marked difference in compactness (InfoC and C in Table~\ref{tab:main}, row sparsity in normalized mutual information heatmaps of \cref{fig:qualitative_isaac_main}) reflects naive Tripod's reluctance to deactivate latents due to its vanilla Hessian penalty leg: shrinking latents increases the curvature of the decoder. In contrast, the invariance to latent scaling built into Tripod's NHP leg enables latent deactivation.

\textbf{Ablation studies on \ourmethod{}.}
We ablate each leg of \ourmethod{} in turn. We take the exact configuration we use for each dataset and either set the corresponding regularization weight to $0$ for ablating NHP or KLM, or change the number of quantized values from $12$ to $12^2$ for ablating FSQ. \ourmethod{} w/o NHP takes a significant hit in all modularity and compactness metrics, concretely demonstrating a beneficial application of the Hessian penalty to autoencoders for the first time. \ourmethod{} w/o KLM suffers a slightly cushioned blow in the same metrics. Interestingly, \ourmethod{} w/ finer quantization incurs the most significant penalty, suggesting that the enforced compression and latent organization afforded by finite scalar quantization underpins \ourmethod{}.

\subsection{Qualitative Results}
For each dataset, we qualitatively compare \ourmethod{} and naive \ourmethod{} in terms of how they decode under latent interventions. For reference, we also visualize the pairwise mutual information heatmap between sources and latents (from which the InfoM and InfoC metrics are calculated). \cref{fig:qualitative_isaac_main} presents this for Isaac3D. Due to space constraints, the rest of this material can be found in \cref{app:qualitative_results}. We find that the quantitative improvement of \ourmethod{} over naive \ourmethod{} is mirrored in how consistently intervening on a particular latent dimension  affects generation.

\section{Related Work} \label{sec:related_work}

\textbf{Disentanglement and identifiability.} The classic problem of independent components analysis (ICA;~\citet{comon1994independent,hyv_arinen2000independent}) can be viewed as the progenitor of disentangled representation learning. Notoriously, disentanglement is theoretically underspecified~\citep{hyv_arinen2000independent,locatello2019challenging} when the data-generating process is nonlinear~\citep{hyv_arinen1999nonlinear}, so disentangling in practice relies heavily on inductive biases. 

\textbf{Architectural inductive biases for disentanglement.} We build on the architectural inductive bias of latent quantization~\citep{hsu2023disentanglement}. \citet{leeb2023structure} demonstrate that restricting different latents to enter the decoding computation graph at different points can enable disentanglement. We view this as specifying a rather specific structure for the latent space (essentially assuming one source variable per hierarchy level) and opt not to use it. 

\textbf{Disentanglement via regularizing autoencoders.} Our work follows in the tradition of applying regularizations on autoencoder architectures to encourage disentanglement. Specifically, the KLM regularization falls into an information-theoretic family of regularization techniques. The simplest example of this is the $\beta$-VAE~\citep{higgins2017beta}, which places a heavier weight on the KL-divergence term in the VAE evidence lower bound. To curb the trade-off between disentanglement and reconstruction, \citet{burgess2017understanding} modify $\beta$-VAE to increase the information capacity of latent representations during training. \citet{chen2018isolating} identify a decomposition of the KL-divergence into $3$ terms and argue that only the total correlation term has a disentangling effect; their proposed model, $\beta$-TCVAE, upweights only this term. Similarly, \citet{kim2018disentangling} also regularize latent multiinformation, but use an auxiliary discriminator and the density-ratio trick to estimate multiinformation. We opt not to use this method due to the unwieldy auxiliary discriminator. \citet{kumar2018variational} investigate a more thorough optimization of the KL-divergence through moment matching. \citet{whittington2023disentanglement} show that biologically-inspired constraints that minimize latent activity and weight energy while promoting latent non-negativity help models learn disentangled representations. 

\textbf{Functional inductive biases on the learned data-generating mapping.} \ourmethod{} also incorporates a functional inductive bias on the latent-to-data mapping. \citet{peebles2020hessian} and \citet{wei2021orthogonal} propose regularizing the derivatives of a GAN's generator to minimize inter-latent dependencies. While the former focuses on the mixed derivatives (off-diagonal elements of the Hessian), the latter regularizes the columns of the Jacobian to be orthogonal. Similarly, \citet{gresele2021independent} mathematically show how a Jacobian column orthogonality criterion can rule out classic indeterminancies in nonlinear independent components analysis. We find the vanishing mixed derivative criterion more conceptually appealing, especially since near-orthgonality becomes trivial in high-dimensional activation and data spaces. In a separate vein, \citet{sorrenson2020disentanglement} and \citet{yang2022volume} investigate the assumption of volume preservation in data generation. Finally, sparsity has seen considerable attention in works pursuing  identifiability~\citep{jing2020implicit,zheng2022identifiability,moran2022identifiable}.

\section{Discussion}
In this work, we meld three previously proposed ideas for disentanglement into \ourmethod{}, a method that makes necessary modifications to existing instantiations of these ideas so that a synergy between the components is realized in practice. These conceptual and technical contributions are validated in our experiments: both the specific set of three inductive biases as well as our modifications are essential to \ourmethod{}'s performance. However, one trade-off is an increased computational footprint due to the use of multiple inductive biases in \ourmethod{}. A profiling study~(\cref{app:profiling_study}) indicates that FSQ and KLM incur negligible overhead, but NHP increases training iteration runtime by a factor of about $2.5$ due to the extra decoder forward passes required for its regularization term. 

Given the sensitivity of \ourmethod{} to the degree of quantization (i.e., of compression) identified in our ablation study, it may be fruitful to study mechanisms to automatically tune or learn this key (hyper)parameter. Naturally, this would need to be unsupervised or highly label-efficient in order to be practically useful. One potential unsupervised tuning procedure would be to begin with a channel capacity that is too low, and to increase it until reconstruction performance starts saturating. The key assumption here is that the true sources are an optimal or near-optimal compression of the data. For the label-efficient setting, it may well be good enough to simply use disentanglement metrics such as InfoMEC. The fixed latent space bounds and lack of learnable codes in our FSQ implementation also enable on-the-fly adaptation of the degree of quantization.

Our empirical evaluation is limited to image datasets because this is the only modality (beyond toy low-dimensional setups) we are aware of for which there exist datasets that (i) obey the nonlinear ICA data generation assumptions and (ii) enable quantitative evaluation via ground-truth source labels that completely explain the data. A priori, we expect \ourmethod{}, instantiated with appropriate architectures, to also be effective for other modalities, e.g. time series or graphs, as no assumptions specific to images are made in the formulation of any of the three \ourmethod{} legs.

Finally, our work demonstrates that a feasible alternative to racking our brains in search of new inductive biases for disentanglement is to revisit previously proposed ideas and refurbish them for use in tandem. Indeed, it may be that the right set of component techniques for disentanglement already exist and are simply waiting to be put together.

\section*{Acknowledgements}
We gratefully acknowledge the developers of open-source software packages that facilitated this research: NumPy~\citep{harris2020array}, 
JAX~\citep{bradbury2018jax},
Equinox~\citep{kidger2021equinox}, 
and
scikit-learn~\citep{pedregosa2011scikit}. 
We also thank Will Dorrell, Stefano Ermon, Yoonho Lee, Allan Zhou, Asher Spector, and anonymous reviewers for helpful discussions and feedback on previous drafts.
KH was funded by a Sequoia Capital Stanford Graduate Fellowship. This research was also supported by ONR grants N00014-22-1-2621 and MURI N00014-22-1-2740.

\section*{Impact Statement}

We view the problem of disentanglement as a manifestation of the desire to have machine learning models experience the world as we humans do. We are thus optimistic that this work and others like it will have a role to play in empowering human decision-making in a world increasingly permeated with such models. Nevertheless, like many other AI technologies, disentanglement has potential negative impacts. Examples include enhanced disinformation dissemination, more invasive personal profiling from behavioral data, and increased automation of sensitive decision-making. Avenues for mitigating such negative outcomes include technical approaches, e.g. using human-in-the-loop rather than fully automated systems, as well as policy considerations, e.g. regulation guidelines for the appropriate deployment of models. Proactive pursuit of such strategies may well be crucial for ensuring the positive broader impact of disentanglement.

\bibliography{references}

\begin{thebibliography}{45}
\providecommand{\natexlab}[1]{#1}
\providecommand{\url}[1]{\texttt{#1}}
\expandafter\ifx\csname urlstyle\endcsname\relax
  \providecommand{\doi}[1]{doi: #1}\else
  \providecommand{\doi}{doi: \begingroup \urlstyle{rm}\Url}\fi

\bibitem[Bengio(2013)]{bengio2013deep}
Bengio, Y.
\newblock Deep learning of representations: looking forward.
\newblock In \emph{Statistical Language and Speech Processing}, 2013.

\bibitem[Bengio et~al.(2013)Bengio, L{\'e}onard, and Courville]{bengio2013estimating}
Bengio, Y., L{\'e}onard, N., and Courville, A.
\newblock Estimating or propagating gradients through stochastic neurons for conditional computation.
\newblock \emph{arXiv preprint arXiv:1308.3432}, 2013.

\bibitem[Bradbury et~al.(2018)Bradbury, Frostig, Hawkins, Johnson, Leary, Maclaurin, Necula, Paszke, Vander{P}las, Wanderman-{M}ilne, and Zhang]{bradbury2018jax}
Bradbury, J., Frostig, R., Hawkins, P., Johnson, M.~J., Leary, C., Maclaurin, D., Necula, G., Paszke, A., Vander{P}las, J., Wanderman-{M}ilne, S., and Zhang, Q.
\newblock {JAX}: composable transformations of {P}ython+{N}um{P}y programs, 2018.
\newblock URL \url{http://github.com/google/jax}.

\bibitem[Burgess \& Kim(2018)Burgess and Kim]{burgess20183d}
Burgess, C. and Kim, H.
\newblock 3{D} shapes dataset, 2018.
\newblock URL \url{https://github.com/deepmind/3dshapes-dataset/}.

\bibitem[Burgess et~al.(2017)Burgess, Higgins, Pal, Matthey, Watters, Desjardins, and Lerchner]{burgess2017understanding}
Burgess, C.~P., Higgins, I., Pal, A., Matthey, L., Watters, N., Desjardins, G., and Lerchner, A.
\newblock Understanding disentangling in $\beta$-{VAE}.
\newblock In \emph{NIPS Workshop on Learning Disentangled Representations}, 2017.

\bibitem[Chen et~al.(2018)Chen, Li, Grosse, and Duvenaud]{chen2018isolating}
Chen, R.~T., Li, X., Grosse, R.~B., and Duvenaud, D.~K.
\newblock Isolating sources of disentanglement in variational autoencoders.
\newblock In \emph{Advances in Neural Information Processing Systems}, 2018.

\bibitem[Comon(1994)]{comon1994independent}
Comon, P.
\newblock Independent component analysis, a new concept?
\newblock \emph{Signal Processing}, 36\penalty0 (3):\penalty0 287--314, 1994.

\bibitem[Dhariwal \& Nichol(2021)Dhariwal and Nichol]{dhariwal2021diffusion}
Dhariwal, P. and Nichol, A.
\newblock Diffusion models beat {GANs} on image synthesis.
\newblock \emph{Advances in Neural Information Processing Systems}, 2021.

\bibitem[Eastwood \& Williams(2018)Eastwood and Williams]{eastwood2018framework}
Eastwood, C. and Williams, C.~K.
\newblock A framework for the quantitative evaluation of disentangled representations.
\newblock In \emph{International Conference on Learning Representations}, 2018.

\bibitem[Gondal et~al.(2019)Gondal, Wuthrich, Miladinovic, Locatello, Breidt, Volchkov, Akpo, Bachem, Sch\"{o}lkopf, and Bauer]{gondal2019transfer}
Gondal, M.~W., Wuthrich, M., Miladinovic, D., Locatello, F., Breidt, M., Volchkov, V., Akpo, J., Bachem, O., Sch\"{o}lkopf, B., and Bauer, S.
\newblock On the transfer of inductive bias from simulation to the real world: a new disentanglement dataset.
\newblock In \emph{Advances in Neural Information Processing Systems}, 2019.

\bibitem[Goodfellow et~al.(2014)Goodfellow, Pouget-Abadie, Mirza, Xu, Warde-Farley, Ozair, Courville, and Bengio]{goodfellow2014generative}
Goodfellow, I., Pouget-Abadie, J., Mirza, M., Xu, B., Warde-Farley, D., Ozair, S., Courville, A., and Bengio, Y.
\newblock Generative adversarial networks.
\newblock In \emph{Advances in Neural Information Processing Systems}, 2014.

\bibitem[Gresele et~al.(2021)Gresele, Von~K{\"u}gelgen, Stimper, Sch{\"o}lkopf, and Besserve]{gresele2021independent}
Gresele, L., Von~K{\"u}gelgen, J., Stimper, V., Sch{\"o}lkopf, B., and Besserve, M.
\newblock Independent mechanism analysis, a new concept?
\newblock \emph{Advances in Neural Information Processing Systems}, 2021.

\bibitem[Harris et~al.(2020)Harris, Millman, Van Der~Walt, Gommers, Virtanen, Cournapeau, Wieser, Taylor, Berg, Smith, et~al.]{harris2020array}
Harris, C.~R., Millman, K.~J., Van Der~Walt, S.~J., Gommers, R., Virtanen, P., Cournapeau, D., Wieser, E., Taylor, J., Berg, S., Smith, N.~J., et~al.
\newblock Array programming with {NumPy}.
\newblock \emph{Nature}, 585\penalty0 (7825):\penalty0 357--362, 2020.

\bibitem[Higgins et~al.(2017)Higgins, Matthey, Pal, Burgess, Glorot, Botvinick, Mohamed, and Lerchner]{higgins2017beta}
Higgins, I., Matthey, L., Pal, A., Burgess, C., Glorot, X., Botvinick, M., Mohamed, S., and Lerchner, A.
\newblock $\beta$-{VAE}: learning basic visual concepts with a constrained variational framework.
\newblock In \emph{International Conference on Learning Representations}, 2017.

\bibitem[Horan et~al.(2021)Horan, Richardson, and Weiss]{horan2021when}
Horan, D., Richardson, E., and Weiss, Y.
\newblock When is unsupervised disentanglement possible?
\newblock In \emph{Advances in Neural Information Processing Systems}, 2021.

\bibitem[Hsu et~al.(2023)Hsu, Dorrell, Whittington, Wu, and Finn]{hsu2023disentanglement}
Hsu, K., Dorrell, W., Whittington, J.~C., Wu, J., and Finn, C.
\newblock Disentanglement via latent quantization.
\newblock \emph{Advances in Neural Information Processing Systems}, 2023.

\bibitem[Hutchinson(1989)]{hutchinson1989stochastic}
Hutchinson, M.~F.
\newblock A stochastic estimator of the trace of the influence matrix for laplacian smoothing splines.
\newblock \emph{Communications in Statistics-Simulation and Computation}, 18\penalty0 (3):\penalty0 1059--1076, 1989.

\bibitem[Hyv{\"a}rinen \& Oja(2000)Hyv{\"a}rinen and Oja]{hyv_arinen2000independent}
Hyv{\"a}rinen, A. and Oja, E.
\newblock Independent component analysis: algorithms and applications.
\newblock \emph{Neural Networks}, 13\penalty0 (4-5):\penalty0 411--430, 2000.

\bibitem[Hyv{\"a}rinen \& Pajunen(1999)Hyv{\"a}rinen and Pajunen]{hyv_arinen1999nonlinear}
Hyv{\"a}rinen, A. and Pajunen, P.
\newblock Nonlinear independent component analysis: existence and uniqueness results.
\newblock \emph{Neural Networks}, 12\penalty0 (3):\penalty0 429--439, 1999.

\bibitem[Jang et~al.(2017)Jang, Gu, and Poole]{jang2017categorical}
Jang, E., Gu, S., and Poole, B.
\newblock Categorical reparameterization with {Gumbel-Softmax}.
\newblock In \emph{International Conference on Learning Representations}, 2017.

\bibitem[Jing et~al.(2020)Jing, Zbontar, et~al.]{jing2020implicit}
Jing, L., Zbontar, J., et~al.
\newblock Implicit rank-minimizing autoencoder.
\newblock In \emph{Advances in Neural Information Processing Systems}, 2020.

\bibitem[Khemakhem et~al.(2020)Khemakhem, Kingma, Monti, and Hyvarinen]{khemakhem2020variational}
Khemakhem, I., Kingma, D., Monti, R., and Hyvarinen, A.
\newblock Variational autoencoders and nonlinear {ICA}: a unifying framework.
\newblock In \emph{International Conference on Artificial Intelligence and Statistics}, 2020.

\bibitem[Kidger \& Garcia(2021)Kidger and Garcia]{kidger2021equinox}
Kidger, P. and Garcia, C.
\newblock Equinox: neural networks in {JAX} via callable {P}y{T}rees and filtered transformations.
\newblock \emph{Differentiable Programming Workshop at Neural Information Processing Systems}, 2021.

\bibitem[Kim \& Mnih(2018)Kim and Mnih]{kim2018disentangling}
Kim, H. and Mnih, A.
\newblock Disentangling by factorising.
\newblock In \emph{International Conference on Machine Learning}, 2018.

\bibitem[Kingma \& Welling(2014)Kingma and Welling]{kingma2014auto}
Kingma, D.~P. and Welling, M.
\newblock Auto-encoding variational {Bayes}.
\newblock In \emph{International Conference on Learning Representations}, 2014.

\bibitem[Kumar et~al.(2018)Kumar, Sattigeri, and Balakrishnan]{kumar2018variational}
Kumar, A., Sattigeri, P., and Balakrishnan, A.
\newblock Variational inference of disentangled latent concepts from unlabeled observations.
\newblock In \emph{International Conference on Learning Representations}, 2018.

\bibitem[Leeb et~al.(2023)Leeb, Lanzillotta, Annadani, Besserve, Bauer, and Sch{\"o}lkopf]{leeb2023structure}
Leeb, F., Lanzillotta, G., Annadani, Y., Besserve, M., Bauer, S., and Sch{\"o}lkopf, B.
\newblock Structure by architecture: structured representations without regularization.
\newblock In \emph{International Conference on Learning Representations}, 2023.

\bibitem[Locatello et~al.(2019)Locatello, Bauer, Lucic, Raetsch, Gelly, Sch{\"o}lkopf, and Bachem]{locatello2019challenging}
Locatello, F., Bauer, S., Lucic, M., Raetsch, G., Gelly, S., Sch{\"o}lkopf, B., and Bachem, O.
\newblock Challenging common assumptions in the unsupervised learning of disentangled representations.
\newblock In \emph{International Conference on Machine Learning}, 2019.

\bibitem[Maddison et~al.(2017)Maddison, Mnih, and Teh]{maddison2017concrete}
Maddison, C.~J., Mnih, A., and Teh, Y.~W.
\newblock The concrete distribution: a continuous relaxation of discrete random variables.
\newblock In \emph{International Conference on Learning Representations}, 2017.

\bibitem[Mentzer et~al.(2024)Mentzer, Minnen, Agustsson, and Tschannen]{mentzer2024finite}
Mentzer, F., Minnen, D., Agustsson, E., and Tschannen, M.
\newblock Finite scalar quantization: {VQ-VAE} made simple.
\newblock In \emph{International Conference on Learning Representations}, 2024.

\bibitem[Moran et~al.(2022)Moran, Sridhar, Wang, and Blei]{moran2022identifiable}
Moran, G.~E., Sridhar, D., Wang, Y., and Blei, D.
\newblock Identifiable deep generative models via sparse decoding.
\newblock \emph{Transactions on Machine Learning Research}, 2022.

\bibitem[Nie(2019)]{nie2019high}
Nie, W.
\newblock High resolution disentanglement datasets, 2019.
\newblock URL \url{https://github.com/NVlabs/High-res-disentanglement-datasets}.

\bibitem[Oord et~al.(2017)Oord, Vinyals, and Kavukcuoglu]{oord2017neural}
Oord, A. v.~d., Vinyals, O., and Kavukcuoglu, K.
\newblock Neural discrete representation learning.
\newblock In \emph{Advances in Neural Information Processing Systems}, 2017.

\bibitem[Pedregosa et~al.(2011)Pedregosa, Varoquaux, Gramfort, Michel, Thirion, Grisel, Blondel, Prettenhofer, Weiss, Dubourg, Vanderplas, Passos, Cournapeau, Brucher, Perrot, and {{\'E}}douard Duchesnay]{pedregosa2011scikit}
Pedregosa, F., Varoquaux, G., Gramfort, A., Michel, V., Thirion, B., Grisel, O., Blondel, M., Prettenhofer, P., Weiss, R., Dubourg, V., Vanderplas, J., Passos, A., Cournapeau, D., Brucher, M., Perrot, M., and {{\'E}}douard Duchesnay.
\newblock Scikit-learn: machine learning in {P}ython.
\newblock \emph{Journal of Machine Learning Research}, 12\penalty0 (85):\penalty0 2825--2830, 2011.

\bibitem[Peebles et~al.(2020)Peebles, Peebles, Zhu, Efros, and Torralba]{peebles2020hessian}
Peebles, W., Peebles, J., Zhu, J.-Y., Efros, A., and Torralba, A.
\newblock The {H}essian penalty: a weak prior for unsupervised disentanglement.
\newblock In \emph{European Conference on Computer Vision}, 2020.

\bibitem[Rudin et~al.(2022)Rudin, Chen, Chen, Huang, Semenova, and Zhong]{rudin2022interpretable}
Rudin, C., Chen, C., Chen, Z., Huang, H., Semenova, L., and Zhong, C.
\newblock Interpretable machine learning: fundamental principles and 10 grand challenges.
\newblock \emph{Statistic Surveys}, 16:\penalty0 1--85, 2022.

\bibitem[Silverman(1998)]{silverman2018density}
Silverman, B.~W.
\newblock \emph{Density estimation for statistics and data analysis}.
\newblock Routledge, 1998.

\bibitem[Sorrenson et~al.(2020)Sorrenson, Rother, and K{\"o}the]{sorrenson2020disentanglement}
Sorrenson, P., Rother, C., and K{\"o}the, U.
\newblock Disentanglement by nonlinear {ICA} with general incompressible-flow networks ({GIN}).
\newblock In \emph{International Conference on Learning Representations}, 2020.

\bibitem[Studen{\`y} \& Vejnarov{\'a}(1998)Studen{\`y} and Vejnarov{\'a}]{studeny1998multiinformation}
Studen{\`y}, M. and Vejnarov{\'a}, J.
\newblock The multiinformation function as a tool for measuring stochastic dependence.
\newblock \emph{Learning in Graphical Models}, pp.\  261--297, 1998.

\bibitem[Wang et~al.(2023)Wang, Xiao, Seyde, Hasani, and Rus]{wang2023measuring}
Wang, T.-H., Xiao, W., Seyde, T., Hasani, R., and Rus, D.
\newblock Measuring interpretability of neural policies of robots with disentangled representation.
\newblock In \emph{Conference on Robot Learning}, 2023.

\bibitem[Wei et~al.(2021)Wei, Shi, Liu, Ji, Gao, Wu, and Zuo]{wei2021orthogonal}
Wei, Y., Shi, Y., Liu, X., Ji, Z., Gao, Y., Wu, Z., and Zuo, W.
\newblock Orthogonal {Jacobian} regularization for unsupervised disentanglement in image generation.
\newblock In \emph{International Conference on Computer Vision}, 2021.

\bibitem[Whittington et~al.(2023)Whittington, Dorrell, Ganguli, and Behrens]{whittington2023disentanglement}
Whittington, J. C.~R., Dorrell, W., Ganguli, S., and Behrens, T.
\newblock Disentanglement with biological constraints: a theory of functional cell types.
\newblock In \emph{International Conference on Learning Representations}, 2023.

\bibitem[Yang et~al.(2022)Yang, Yang, Sun, Zhang, Zhang, Li, and Yan]{yang2022volume}
Yang, X., Yang, Y., Sun, J., Zhang, X., Zhang, S., Li, Z., and Yan, J.
\newblock Nonlinear {ICA} using volume-preserving transformations.
\newblock In \emph{International Conference on Learning Representations}, 2022.

\bibitem[Zheng \& Lapata(2022)Zheng and Lapata]{zheng2022disentangled}
Zheng, H. and Lapata, M.
\newblock Disentangled sequence to sequence learning for compositional generalization.
\newblock In \emph{Proceedings of the 60th Annual Meeting of the Association for Computational Linguistics (Volume 1: Long Papers)}, 2022.

\bibitem[Zheng et~al.(2022)Zheng, Ng, and Zhang]{zheng2022identifiability}
Zheng, Y., Ng, I., and Zhang, K.
\newblock On the identifiability of nonlinear {ICA}: sparsity and beyond.
\newblock In \emph{Advances in Neural Information Processing Systems}, 2022.

\end{thebibliography}
\bibliographystyle{icml2024}

\newpage
\appendix
\onecolumn


\section{Proofs}
\subsection{Proof of \cref{prop:hp_vs_nhp}} \label{app:hp_vs_nhp_proof}
\textbf{Proposition 3.1.} \textit{The Hessian penalty
    \begin{equation}
        \sum_{j_1\neq j_2} \left(H^{[k]}_{j_1 j_2}\right)^2
    \end{equation}
can be reduced by scaling down $\decoder^{[k]}$ or scaling up any $z_j, j \in [\numof{\latent}]$, and vice versa. In contrast, the normalized Hessian penalty
    \begin{equation}
        \frac{\sum_{j_1\neq j_2} \left(H^{[k]}_{j_1 j_2} \sigma_{j_1} \sigma_{j_2}\right)^2}{\sum_{j_1,j_2} \left(H^{[k]}_{j_1 j_2} \sigma_{j_1} \sigma_{j_2}\right)^2}  
    \end{equation}
is invariant to the scaling of ${\decoder}^{[k]}$ and $\latent_j \ \forall j \in [\numof{\latent}]$.}

\begin{proof}
First, we prove the statement about the original Hessian penalty. Let $s \in (0, 1)$. Then, if $\decoder'^{[k]} = s\decoder^{[k]}$, 
\begin{align*}
    \sum_{j_1\neq j_2} \left(H'^{[k]}_{j_1 j_2}\right)^2 = \sum_{j_1 \neq j_2}\left( \frac{\partial}{\partial z_{j_1}}\frac{\partial \decoder'^{[k]}}{\partial z_{j_2}} \right)^2 
    = \sum_{j_1 \neq j_2}\left( \frac{\partial}{\partial z_{j_1}}\frac{\partial \decoder'^{[k]}}{\partial \decoder^{[k]}}\frac{\partial \decoder^{[k]}}{\partial z_{j_2}} \right)^2 
    = \sum_{j_1 \neq j_2}\left( \frac{\partial}{\partial z_{j_1}}\frac{\partial \decoder^{[k]}}{\partial z_{j_2}} \right)^2 s^2
    < \sum_{j_1\neq j_2} \left(H^{[k]}_{j_1 j_2}\right)^2.
\end{align*}
Similarly, if $z_i' = s_i z_i$ with $s_i \geq 1$ and $s_l > 1$ for at least one $l \in [n_z]$, then
\begin{align*}
    \sum_{j_1\neq j_2} \left(H'^{[k]}_{j_1 j_2}\right)^2 = \sum_{j_1 \neq j_2}\left( \frac{\partial}{\partial z_{j_1}'}\frac{\partial \decoder^{[k]}}{\partial z_{j_2}'} \right)^2 
    = \sum_{j_1 \neq j_2}\left( \frac{\partial}{\partial z_{j_1}} \frac{\partial z_{j_1}}{\partial z_{j_1}'} \frac{\partial \decoder^{[k]}}{\partial z_{j_2}}\frac{\partial z_{j_2}}{\partial z_{j_2}'} \right)^2 
    = \sum_{j_1 \neq j_2}\left( \frac{\partial}{\partial z_{j_1}}\frac{\partial \decoder^{[k]}}{\partial z_{j_2}}  \frac{1}{s_{j_1}s_{j_2}}\right)^2
    < \sum_{j_1\neq j_2} \left(H^{[k]}_{j_1 j_2}\right)^2.
\end{align*}
To show that the normalized Hessian penalty is invariant to scaling of $\decoder^{[k]}$, suppose we scale $\decoder^{[k]}$ to be $\decoder'^{[k]} = \alpha \decoder^{[k]}$. Then, the numerator becomes
\begin{align*}
    \sum_{j_1 \neq j_2} \left(H'^{[k]}_{j_1 j_2}\sigma_{j_1}\sigma_{j_2}\right)^2 = \sum_{j_1 \neq j_2} \left(\frac{\partial}{\partial z_{j_1}}\frac{\partial \decoder'^{[k]}}{\partial z_{j_2}} \sigma_{j_1} \sigma_{j_2}\right)^2
    = \sum_{j_1 \neq j_2} \left(\frac{\partial}{\partial z_{j_1}}\frac{\partial \decoder'^{[k]}}{\partial \decoder^{[k]}}\frac{\partial \decoder^{[k]}}{\partial z_{j_2} } \sigma_{j_1} \sigma_{j_2} \right)^2 
    = \alpha^2 \sum_{j_1 \neq j_2} \left(H^{[k]}_{j_1 j_2} \sigma_{j_1}\sigma_{j_2}\right)^2.
\end{align*}
Similarly, the denominator becomes $\sum_{j_1, j_2} \left(H'^{[k]}_{j_1 j_2}\sigma_{j_1}\sigma_{j_2}\right)^2 = \alpha^2 \sum_{j_1, j_2} \left(H^{[k]}_{j_1 j_2} \sigma_{j_1}\sigma_{j_2}\right)^2$, allowing us to write 
\begin{align*}
    \frac{\sum_{j_1\neq j_2} \left(H'^{[k]}_{j_1 j_2}\sigma_{j_1}\sigma_{j_2}\right)^2}{\sum_{j_1,j_2} \left(H'^{[k]}_{j_1 j_2}\sigma_{j_1}\sigma_{j_2}\right)^2} = \frac{\sum_{j_1\neq j_2} \left(H^{[k]}_{j_1 j_2} \sigma_{j_1} \sigma_{j_2}\right)^2}{\sum_{j_1,j_2} \left(H^{[k]}_{j_1 j_2} \sigma_{j_1} \sigma_{j_2}\right)^2}
\end{align*}
Now, we verify that the normalized Hessian penalty is invariant to any scaling of the inputs. If $z_{j_1}$ is scaled by $s_1$ so that $z_{j_1}' = s_1z_{j_1}$ and $z_{j_2}$ is scaled by $s_2$ so that $z_{j_2}' = s_2z_{j_2}$, then each term (in both the numerator and the denominator) becomes 
    \begin{align*}
        \left(H'^{[k]}_{j_1 j_2} \sigma_{j_1}'\sigma_{j_2}'\right)^2 &= \left(\frac{\partial}{\partial z_{j_1}'}\frac{\partial \decoder^{[k]}}{\partial z_{j_2}'} \sigma_{j_1}' \sigma_{j_2}' \right)^2 \\  
        &= \left( \frac{\partial}{\partial z_{j_1}} \frac{\partial z_{j_1}}{\partial z_{j_1}'} 
 \left(\frac{\partial \decoder^{[k]}}{\partial z_{j_2}}\frac{\partial z_{j_2}}{\partial z_{j_2}'}\right) \sigma_{j_1}' \sigma_{j_2}' \right)^2 \\  
        &= \left( \frac{1}{s_1} \frac{\partial}{\partial z_{j_1}}\left(\frac{1}{s_2} \frac{\partial \decoder^{[k]}}{\partial z_{j_2}}\right) s_1\sigma_{j_1} s_2\sigma_{j_2} \right)^2 \\  
        &= \left( \frac{\partial}{\partial z_{j_1}}\frac{\partial \decoder^{[k]}}{\partial z_{j_2}}\sigma_{j_1} \sigma_{j_2} \right)^2 \\  
        &= \left(H^{[k]}_{j_1 j_2} \sigma_{j_1}\sigma_{j_2}\right)^2,
    \end{align*}
    and the normalized Hessian penalty remains invariant.    
\end{proof}

\subsection{Proof of \cref{prop:nhp_calculation}} \label{app:nhp_calculation_proof}
\textbf{Proposition 3.2.} \textit{Let $v$ and $w$ be random vectors where $v_j \sim \operatorname{Rademacher}(\sigma_j)$ and $w_j \sim \mathcal{N}(0, \sigma_j^2)$. Then the normalized Hessian penalty can be computed as}
    \begin{equation} 
        \frac{\sum_{j_1 \neq j_2} \left(H^{[k]}_{j_1 j_2} \sigma_{j_1} \sigma_{j_2}\right)^2} { \sum_{j_1, j_2} \left(H^{[k]}_{j_1 j_2} \sigma_{j_1} \sigma_{j_2}\right)^2} = \frac{\Var\left[v^T H^{[k]} v\right]}{\Var\left[w^T H^{[k]} w\right]} .
    \end{equation} 

\begin{proof}

For the numerator, we first write $\Var\left[ v^\top H^{[k]} v \right] = \Var\left[v^\top H^{[k]} v - \E\left[v^\top H^{[k]} v \right]\right]$. Now, 
\begin{align*}
v^\top H^{[k]} v - \E\left[v^\top H^{[k]} v \right] &= \sum_{j_1,j_2} v_{j_1} H^{[k]}_{j_1 j_2} v_{j_2} - \E\left[v^\top H^{[k]} v \right] \\ 
&= \sum_{j_1,j_2} v_{j_1} H^{[k]}_{j_1 j_2} v_{j_2} - \E\left[\Tr\left(v^\top H^{[k]} v\right)\right] \\ 
&= \sum_{j_1,j_2} v_{j_1} H^{[k]}_{j_1 j_2} v_{j_2} - \E\left[\Tr\left(H^{[k]}vv^T\right)\right] \\
&= \sum_{j_1,j_2} v_{j_1} H^{[k]}_{j_1 j_2} v_{j_2} - \Tr\left(H^{[k]}\E\left[vv^T\right]\right) \\
&= \sum_{j_1,j_2} v_{j_1} H^{[k]}_{j_1 j_2} v_{j_2} - \sum_j H^{[k]}_{jj}\sigma_j^2\\
&= \sum_{j_1,j_2} v_{j_1} H^{[k]}_{j_1 j_2} v_{j_2} -\sum_j H^{[k]}_{jj}v_j^2 + \sum_j H^{[k]}_{jj}v_j^2 - \sum_j H^{[k]}_{jj}\sigma_j^2\\
&= \sum_{j_1\neq j_2} v_{j_1} H^{[k]}_{j_1 j_2} v_{j_2} + \sum_j H^{[k]}_{jj}(v_j^2 - \sigma_j^2)\\
&= \sum_{j_1\neq j_2} v_{j_1} H^{[k]}_{j_1 j_2} v_{j_2} + \sum_j H^{[k]}_{jj}(\sigma_j^2 - \sigma_j^2) \qquad \because v_j \sim \operatorname{Rademacher}(\sigma_j)\\
&= \sum_{j_1\neq j_2} v_{j_1} H^{[k]}_{j_1 j_2} v_{j_2}
\end{align*}
Taking the variance, we have: 
\begin{align*}
    \Var\left[\sum_{j_1\neq j_2} v_{j_1} H^{[k]}_{j_1 j_2} v_{j_2}\right]
    &= \Var\left[2\sum_{j_1>j_2}v_{j_1} H^{[k]}_{j_1 j_2} v_{j_2}\right] &\because H^{[k]} \text{ is symmetric} \\
    &= 4\sum_{j_1>j_2}\left(H^{[k]}_{j_1 j_2}\right)^2\Var\left[v_{j_1} v_{j_2} \right]\\
    &= 4\sum_{j_1>j_2}\left(H^{[k]}_{j_1 j_2}\right)^2\sigma_{j_1}^2\sigma_{j_2}^2\\
    &= 2\sum_{j_1\neq j_2}\left(H^{[k]}_{j_1 j_2} \sigma_{j_1} \sigma_{j_2}\right)^2 &\because H^{[k]} \text{ is symmetric} \\
\end{align*}
For the denominator, we have
\begin{align*}
    \Var\left[w^\top H^{[k]} w\right] &= \Var\left[\sum_{j_1\neq j_2}w_{j_1}H^{[k]}_{j_1 j_2}w_{j_2} + \sum_j w_j^2H^{[k]}_{jj}\right] \\
    &= \Var\left[\sum_{j_1}\left(2\sum_{j_1 > j_2} w_{j_1}H^{[k]}_{j_1 j_2}w_{j_2} + w_{j_1}^2H^{[k]}_{j_1 j_1}\right)\right] \\
    &= \sum_{j_1}\Var\left[\left(2\sum_{j_1 > j_2} w_{j_1}H^{[k]}_{j_1 j_2}w_{j_2} + w_{j_1}^2H^{[k]}_{j_1 j_1}\right)\right] \\ &\hspace{-5em} + \sum_{j_1 \neq j_1'}\Cov\left(\left(2\sum_{j_1 > j_2} w_{j_1}H^{[k]}_{j_1 j_2}w_{j_2} + w_{j_1}^2H^{[k]}_{j_1 j_1}\right),  \left(2\sum_{j_1' > j_2'} w_{j_1'}H^{[k]}_{j_1' j_2'}w_{j_2'} + w_{j_1'}^2H^{[k]}_{j_1' j_1'}\right)\right)
\end{align*}
The covariance terms vanish. To see this, note that 
\begin{align*}
    \Cov\left(w_{j_1} H^{[k]}_{j_1 j_2} w_{j_2}, w_{j_1'} H^{[k]}_{j_1' j_2'} w_{j_2'}\right) &= \E\left[w_{j_1} H^{[k]}_{j_1 j_2} w_{j_2} w_{j_1'} H^{[k]}_{j_1' j_2'}w _{j_2'} \right] - \E\left[w_{j_1}H^{[k]}_{j_1 j_2}w_{j_2}\right]\E\left[w_{j_1'} H^{[k]}_{j_1' j_2'} w_{j_2'} \right] \\
    &= \E\left[w_{j_1}\right] \E\left[H^{[k]}_{j_1 j_2} w_{j_2} w_{j_1'} H^{[k]}_{j_1' j_2'}w _{j_2'} \right] - \E\left[w_{j_1}\right] \E\left[H^{[k]}_{j_1 j_2}w_{j_2}\right]\E\left[w_{j_1'}\right]\E\left[ H^{[k]}_{j_1' j_2'} w_{j_2'} \right] \\
    &= 0,
\end{align*}
where we leverage that either $j_1 \neq j'_2$ or $j_2 \neq j_1'$ must hold else the two terms are identical, and write the case of the former without loss of generality.
Similarly,  
\begin{align*}
    \Cov\left(w_{j_1} H^{[k]}_{j_1 j_2} w_{j_2}, w_{j_1'}^2 H^{[k]}_{j_1' j_1'}\right) = \Cov\left(w_{j_1} H^{[k]}_{j_1 j_2} w_{j_2}, w_{j_1}^2 H^{[k]}_{j_1 j_1}\right) = 0
\end{align*}
since in each of these expressions there is at least one normal variable independent of all other variables that has not been raised to a power greater than 1. 
Lastly, 
\begin{align*}
    \Cov\left(w_{j_1}^2H^{[k]}_{j_1j_1}, w_{j_1'}^2H^{[k]}_{j_1'j_1'}\right) &= \E\left[w_{j_1}^2H^{[k]}_{j_1j_1}w_{j_1'}^2H^{[k]}_{j_1'j_1'}\right] - \E\left[w_{j_1'}^2H^{[k]}_{j_1'j_1'}\right]\E\left[w_{j_1}^2H^{[k]}_{j_1j_1}\right] \\
    &= H^{[k]}_{j_1j_1}H^{[k]}_{j_1'j_1'}\sigma_{j_1}^2\sigma_{j_1'}^2 - H^{[k]}_{j_1j_1}H^{[k]}_{j_1'j_1'}\sigma_{j_1}^2\sigma_{j_1'}^2 \\
    &= 0.
\end{align*} 
Since the covariance terms vanish, we have
\begin{align*}
    \Var\left[w^\top H^{[k]} w\right] 
    &= \Var\left[\sum_{j_1}\left(2\sum_{j_1 > j_2} w_{j_1}H^{[k]}_{j_1 j_2}w_{j_2} + w_{j_1}^2H^{[k]}_{j_1 j_1}\right)\right] \\
    &= \sum_{j_1}\Var\left[2\sum_{j_1 > j_2} w_{j_1}H^{[k]}_{j_1 j_2}w_{j_2} + w_{j_1}^2H^{[k]}_{j_1 j_1}\right].
\end{align*}
Now, we expand: 
\begin{align*}
    &\sum_{j_1}\left(\Var \left[2 \sum_{j_1 > j_2} w_{j_1}H^{[k]}_{j_1 j_2}w_{j_2} + w_{j_1}^2H^{[k]}_{j_1 j_1}\right]\right) \\= &\sum_{j_1} \left( \Var\left[2\sum_{j_1>j_2} w_{j_1}H^{[k]}_{j_1 j_2}w_{j_2}\right] + \Var\left[w_{j_1}^2H^{[k]}_{j_1 j_1}\right] + 2\Cov\left(2\sum_{j_1>j_2} w_{j_1}H^{[k]}_{jj}w_{j_2}, w_{j_1}^2H^{[k]}_{j_1 j_1}\right)\right). 
\end{align*}
However, for all $j_1$, 
\begin{align*}
    2\Cov\left(2\sum_{j_1>j_2} w_{j_1}H^{[k]}_{jj}w_{j_2}, w_{j_1}^2H^{[k]}_{j_1 j_1}\right) &= \E\left[2w_{j_1}^3H^{[k]}_{j_1 j_1}\sum_{j_1>j_2}H^{[k]}_{j_1 j_2}w_{j_2}\right] - \E\left[2w_{j_1}\sum_{j_1>j_2}H^{[k]}_{j_1 j_2}w_{j_2}\right]\E\left[w_{j_1}^2H^{[k]}_{j_1 j_1}\right] \\ 
    &= 2H^{[k]}_{j_1 j_1}\sum_{j_1>j_2}H^{[k]}_{j_1 j_2}\E\left[w_{j_1}^3w_{j_2}\right] - 2H^{[k]}_{j_1 j_1}\sum_{j_1>j_2}H^{[k]}_{j_1 j_2}\E\left[w_{j_1}w_{j_2}\right]\E\left[w_{j_1}^2\right] \\
    &= 2H^{[k]}_{j_1 j_1}\sum_{j_1>j_2}H^{[k]}_{j_1 j_2}\E\left[w_{j_1}^3\right]\E\left[w_{j_2}\right] - 2H^{[k]}_{j_1 j_1}\sum_{j_1>j_2}H^{[k]}_{j_1 j_2}\E\left[w_{j_1}\right]\E\left[w_{j_2}\right]\E\left[w_{j_1}^2\right] \\
    &= 0.
\end{align*}
Therefore, 
\begin{align*}
    \Var\left[w^\top H^{[k]} w\right] &= \sum_{j_1} \left( \Var\left[2\sum_{j_1 > j_2} w_{j_1}H^{[k]}_{j_1 j_2}w_{j_2}\right] +  \Var\left[w_{j_1}^2H^{[k]}_{j_1 j_1}\right]\right) \\
    &= \sum_{j_1} \left(4\sum_{j_1>j_2}\left(H^{[k]}_{j_1 j_2}\right)^2\Var\left[w_{j_1}w_{j_2}\right] + \left(H^{[k]}_{j_1 j_1}\right)^2\Var\left[w_{j_1}^2\right]\right)\\
    &= \sum_{j_1} \left(4\sum_{j_1>j_2}\left(H^{[k]}_{j_1 j_2}\right)^2 \sigma_{j_1}^2 \sigma_{j_1}^2 + 2\left(H^{[k]}_{jj}\right)^2 \sigma_{j_1}^2\right)\\
    &= 2\sum_{j_1,j_2}\left(H^{[k]}_{j_1 j_2}\right)^2 \sigma_{j_1}^2 \sigma_j^2
\end{align*}
where we used the fact that the variance of product of two independent normal variables centered at 0 is the product of their variances and the variance of the square of a normal variable is twice its variance squared. 

Taking the simplified forms of the numerator and denominator, we obtain 
\begin{align*}
    \frac{ \Var\left[ v^\top H^{[k]} v \right]}{\Var\left[w^\top H^{[k]} w\right]} &= \frac{2\sum_{j_1\neq j_2} \left(H^{[k]}_{j_1 j_2}\sigma_{j_1} \sigma_{j_2} \right)^2}{2\sum_{j_1,j_2} \left(H^{[k]}_{j_1 j_2} \sigma_{j_1} \sigma_{j_2}\right)^2} = \frac{\sum_{j_1 \neq j_2} \left(H^{[k]}_{j_1 j_2} \sigma_{j_1} \sigma_{j_2}\right)^2} { \sum_{j_1, j_2} \left(H^{[k]}_{j_1 j_2} \sigma_{j_1} \sigma_{j_2}\right)^2}
\end{align*}
\end{proof}

\newpage

\section{Experimental Details} \label{app:experiment_details}

This section contains details on the experiments conducted in this work.

\subsection{Datasets} \label{app:datasets}

\begin{figure}[ht]
\centering
\begin{subfigure}{0.5\textwidth}
\centering
\includegraphics[width=0.9\linewidth]{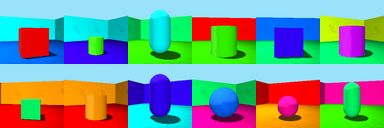}
\caption{Shapes3D}
\end{subfigure}%
\begin{subfigure}{0.5\textwidth}
\centering
\includegraphics[width=0.9\linewidth]{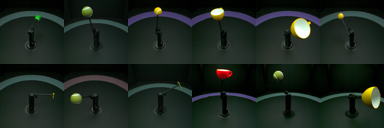}
\caption{MPI3D}
\end{subfigure}

\vspace{10pt}

\begin{subfigure}{0.5\textwidth}
\centering
\includegraphics[width=0.9\linewidth]{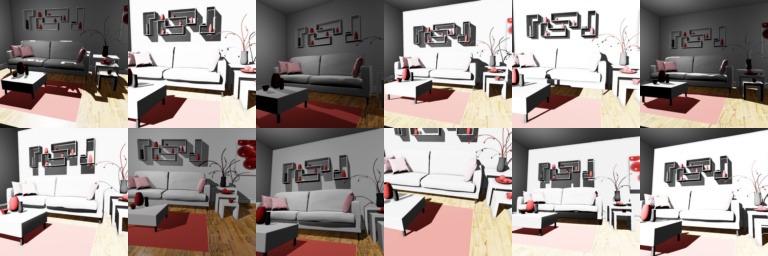}
\caption{Falcor3D}
\end{subfigure}%
\begin{subfigure}{0.5\textwidth}
\centering
\includegraphics[width=0.9\linewidth]{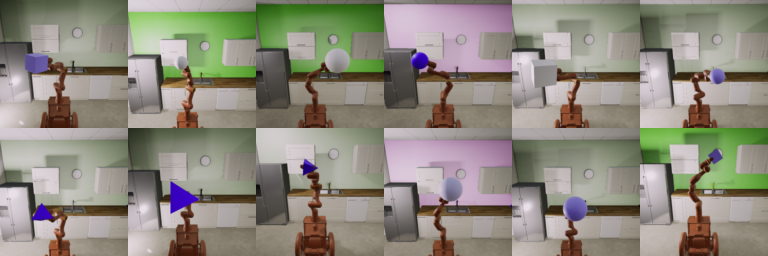}
\caption{Isaac3D}
\end{subfigure}
\caption{Random data samples from each dataset.}
\label{fig:dataset_samples}
\end{figure}

\begin{table}[H]
    \caption{Summary of datasets used for empirical evaluation.}
    \label{tab:datasets}
    \centering
    \small
    \begin{tabular}{lccc}
        \toprule
            dataset & $\numof{\source}$ & $|\dataset|$ & PSNR threshold (dB) for $64 \times 64$ \\
        \midrule
            Shapes3D~\citep{burgess20183d} & $6$ & $\num{480000}$ & 38 \\
            MPI3D~\citep{gondal2019transfer} & $7$ & $\num{460800}$ & 42 \\
            Falcor3D~\citep{nie2019high} & $7$ & $\num{233280}$ & 34 \\
            Isaac3D~\citep{nie2019high} & $9$ & $\num{737280}$ & 40 \\
        \bottomrule
  \end{tabular}
\end{table}

\begin{table}[H]
    \caption{Dataset sources.} \label{tab:sources}
    \centering
    \small
    \begin{tabular}{rlrlrlrlr}
        \toprule
             & \multicolumn{2}{c}{Shapes3D} & \multicolumn{2}{c}{MPI3D} & \multicolumn{2}{c}{Falcor3D} & \multicolumn{2}{c}{Isaac3D} \\
            index & description & values & description & values & description & values & description & values \\
        \midrule
            0 & floor color & 10 & object color & 4 & lighting intensity & 5 & object shape & 3 \\
            1 & object color & 10 & object shape & 4 & lighting x & 6 & robot x & 8 \\
            2 & camera orientation & 10 & object size & 2 & lighting y & 6 & robot y & 5 \\
            3 & object scale & 8 & camera height & 3 & lighting z & 6 & camera height & 4 \\
            4 & object shape & 4 & background color & 3 & camera x & 6 & object scale & 4 \\
            5 & wall color & 15 & robot x & 40 & camera y & 6 & lighting intensity & 4 \\
            6 & & & robot y & 40 & camera z & 6 & lighting direction & 6 \\
            7 & & & & & & & object color & 4 \\
            8 & & & & & & & wall color & 4 \\
        \bottomrule
    \end{tabular}
\end{table}

\newpage
\subsection{Hyperparameters}    \label{app:hyperparameters}

This section specifies fixed and tuned hyperparameters for all methods considered.

\begin{table}[H]
  \caption{Fixed hyperparameters for all autoencoder variants.}
  \label{tab:ae_fixed}
  \centering
  \begin{tabular}{ll}
    \toprule
        hyperparameter & value \\
    \midrule
        number of latents $n_z$ & $2 \numof{\source}$ \\
        AdamW learning rate & ${1 \times 10^{-3}}$ \\
        AdamW $\beta_1$ & $0.9$ \\
        AdamW $\beta_2$ & $0.99$ \\
        AdamW updates & $\leq 2 \times 10^{5}$ \\
        batch size & 64 \\
    \bottomrule
  \end{tabular}
\end{table}

\begin{table}[H]
  \caption{Key regularization hyperparameter tuning done for each autoencoder}
  \label{tab:hyperparameter_tuning}
  \centering
  \begin{tabular}{llll}
    \toprule
        method & hyperparameter & values \\
    \midrule
        $\beta$-TCVAE & $\beta = \weight{total correlation}$ & $[1, 2, 3, 5, 10]$ \\
        QLAE & $\text{weight decay}$ & $[1 \times 10^{-8}, 1 \times 10^{-6}, 1 \times 10^{-4}, 1 \times 10^{-2}, 1]$ \\
        \ourmethod{} (naive) & $\weight{vanilla Hessian penalty}$ & $[0, 1 \times 10^{-10}, 1 \times 10^{-8}, 1 \times 10^{-6}, 1 \times 10^{-4}, 1 \times 10^{-2}]$ \\
        {\color{white}\ourmethod{} (naive)} & $\weight{latent multiinformation}$ & $[0, 1 \times 10^{-10}, 1 \times 10^{-8}, 1 \times 10^{-6}, 1 \times 10^{-4}, 1 \times 10^{-2}]$ \\
        \ourmethod{} & $\lambda_{\text{NHP}}$ & $[0, 1 \times 10^{-10}, 1 \times 10^{-8}, 1 \times 10^{-6}, 1 \times 10^{-4}, 1 \times 10^{-2}]$ \\ 
        {\color{white}\ourmethod{}}  & $\lambda_{\text{KLM}}$ & $[0, 1 \times 10^{-10}, 1 \times 10^{-8}, 1 \times 10^{-6}, 1 \times 10^{-4}, 1 \times 10^{-2}]$ \\
    \bottomrule
  \end{tabular}
\end{table}

\newpage

\section{Qualitative Results} \label{app:qualitative_results}
We qualitatively compare \ourmethod{} and naive \ourmethod{} on each dataset. In each column, we encode an image and visualize the effect of intervening on a single latent on decoding by varying its value in a linear interpolation in that latent's range. Below, we also provide a normalized mutual information heatmap that acts as an ``answer key'' to what the qualitative change in a column should be. Red latents are inactive and corresponding columns are removed from the latent intervention visualizations.

\begin{figure}[H]
    {\scriptsize
    \begin{subfigure}{0.48\textwidth}
        \begin{tabularx}{0.99\textwidth}{*{12}{>{\centering\arraybackslash}X}}
            $z_0$ & $z_1$ & $z_2$ & $z_3$ & $z_4$ & $z_5$ & $z_6$ & \textcolor{red}{$z_7$} & $z_8$ & $z_9$ & \textcolor{red}{$z_{10}$} & $z_{11}$ \\
        \end{tabularx}
    \end{subfigure}
    \hfill
    \begin{subfigure}{0.48\textwidth}
        \begin{tabularx}{0.99\textwidth}{*{12}{>{\centering\arraybackslash}X}}
            $z_0$ & $z_1$ & $z_2$ & $z_3$ & $z_4$ & $z_5$ & $z_6$ & \textcolor{black}{$z_7$} & $z_8$ & $z_9$ & \textcolor{black}{$z_{10}$} & $z_{11}$ \\
        \end{tabularx}
    \end{subfigure}
    }
    \begin{subfigure}{0.48\textwidth}
        \includegraphics[width=1.0\linewidth]{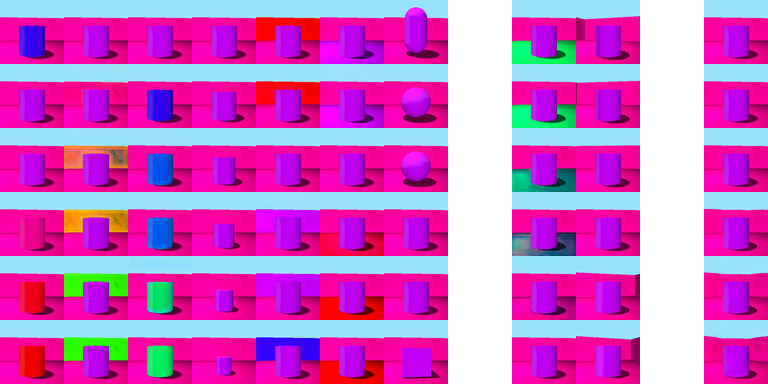}
    \end{subfigure}
    \hfill
    \begin{subfigure}{0.48\textwidth}
        \includegraphics[width=1.0\linewidth]{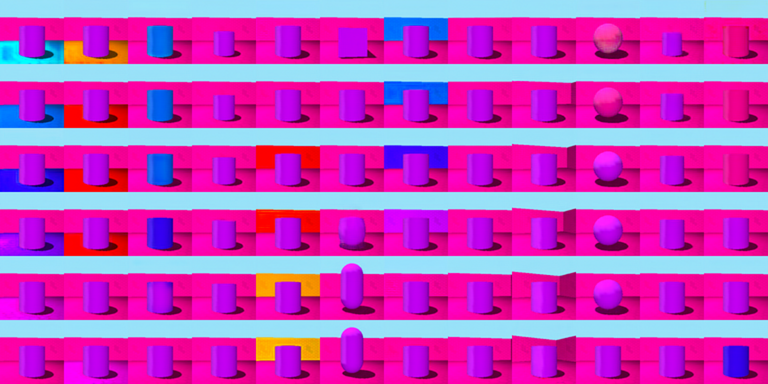}
    \end{subfigure}
    \begin{subfigure}{0.48\textwidth}
        \includegraphics[width=1.0\linewidth]{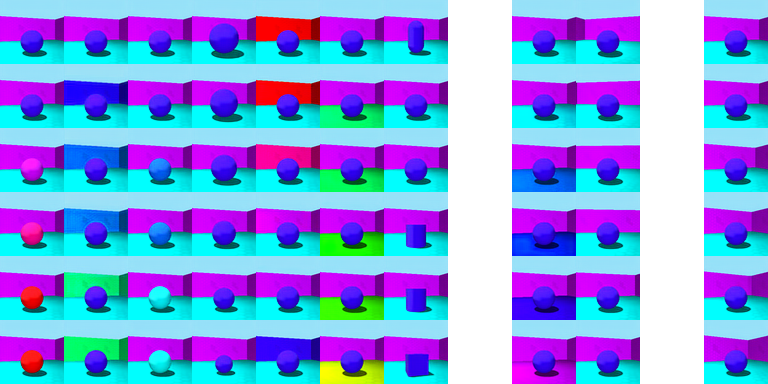}
    \end{subfigure}
    \hfill
    \begin{subfigure}{0.48\textwidth}
        \includegraphics[width=1.0\linewidth]{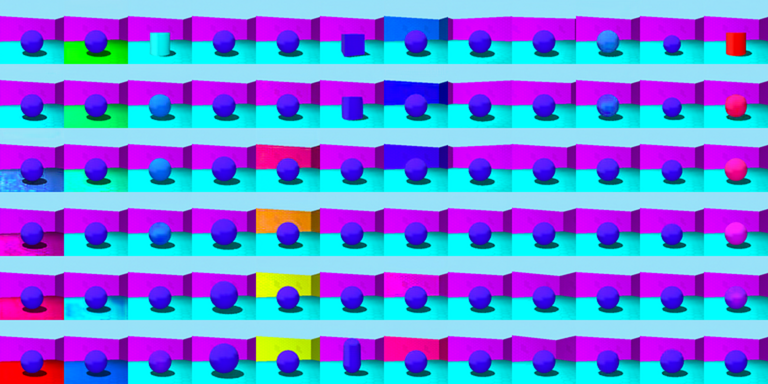}
    \end{subfigure}
    \begin{subfigure}{0.48\textwidth}
        \includegraphics[width=1.0\linewidth]{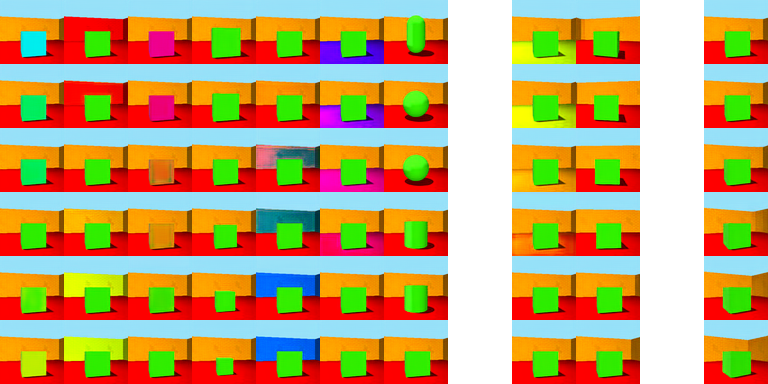}
    \end{subfigure}
    \hfill
    \begin{subfigure}{0.48\textwidth}
        \includegraphics[width=1.0\linewidth]{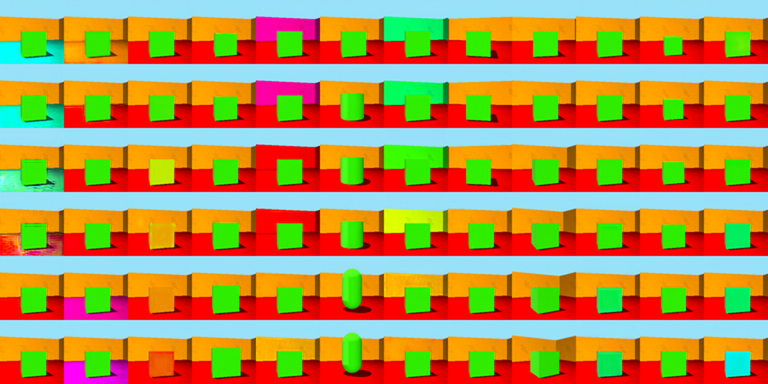}
    \end{subfigure}
    \begin{subfigure}{0.48\textwidth}
        \centering
        \includegraphics[width=1.0\linewidth]{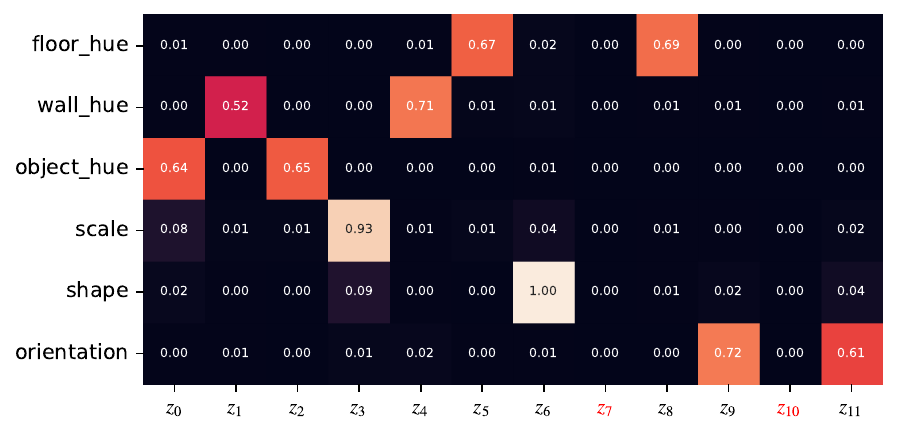}
        \caption{\ourmethod{} }
    \end{subfigure}
    \hfill
    \begin{subfigure}{0.48\textwidth}
        \centering
        \includegraphics[width=1.0\linewidth]{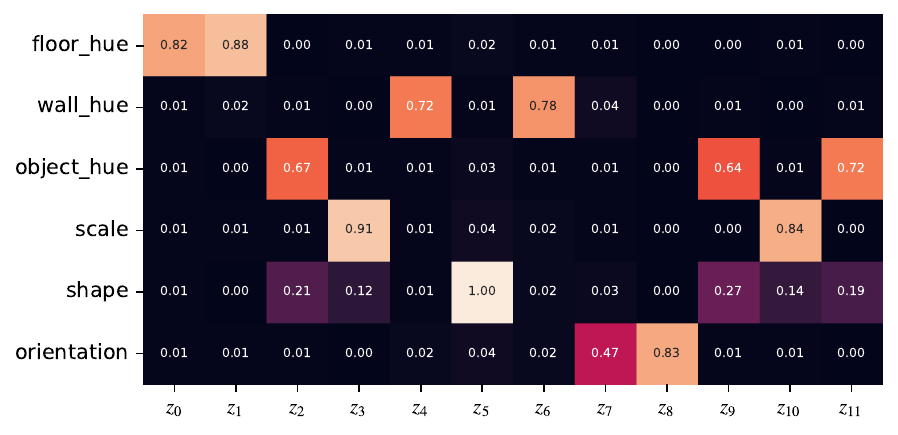}
        \caption{Naive \ourmethod{} }
    \end{subfigure}
    \caption{\ourmethod{} and naive \ourmethod{} decoded latent interventions and normalized mutual information heatmaps on Shapes3D.}
\end{figure}

\begin{figure}[H]
    {\scriptsize
    \begin{subfigure}{0.48\textwidth}
        \begin{tabularx}{0.978\textwidth}{*{14}{>{\centering\arraybackslash}X}}
            $z_0$ & $z_1$ & $z_2$ & $z_3$ & $z_4$ & $z_5$ & $z_6$ & $z_7$ & $z_8$ & $z_9$ & $z_{10}$ & $z_{11}$ & $z_{12}$ & $z_{13}$ \\
        \end{tabularx}
    \end{subfigure}
    \hfill
    \begin{subfigure}{0.48\textwidth}
        \begin{tabularx}{0.978\textwidth}{*{14}{>{\centering\arraybackslash}X}}
            $z_0$ & $z_1$ & $z_2$ & $z_3$ & $z_4$ & $z_5$ & $z_6$ & $z_7$ & $z_8$ & $z_9$ & $z_{10}$ & $z_{11}$ & $z_{12}$ & $z_{13}$ \\
        \end{tabularx}
    \end{subfigure}
    }
    \begin{subfigure}{0.48\textwidth}
        \includegraphics[width=1.0\linewidth]{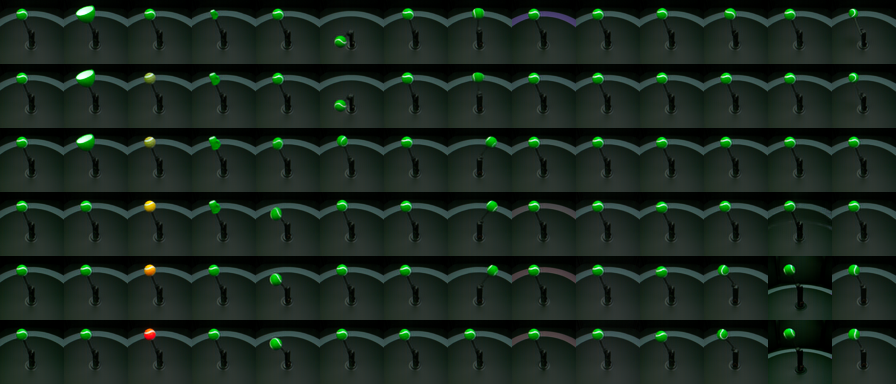}
    \end{subfigure}
    \hfill
    \begin{subfigure}{0.48\textwidth}
        \includegraphics[width=1.0\linewidth]{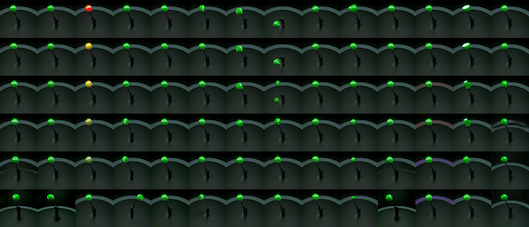}
    \end{subfigure}
    \begin{subfigure}{0.48\textwidth}
        \includegraphics[width=1.0\linewidth]{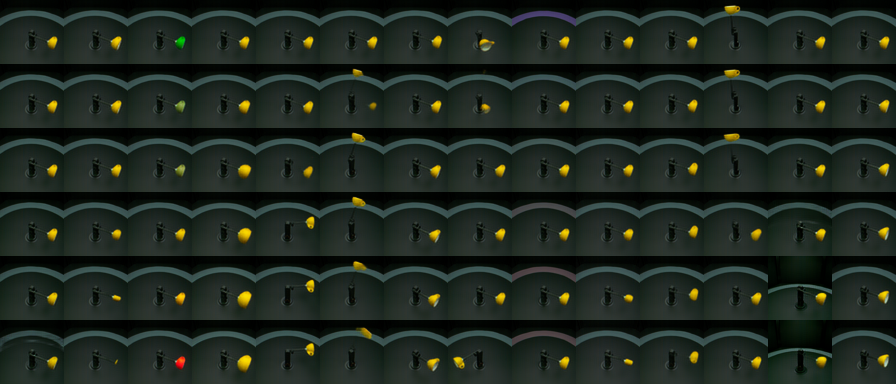}
    \end{subfigure}
    \hfill
    \begin{subfigure}{0.48\textwidth}
        \includegraphics[width=1.0\linewidth]{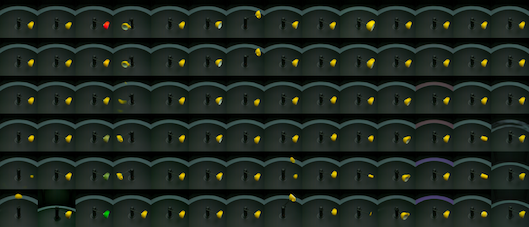}
    \end{subfigure}
    \begin{subfigure}{0.48\textwidth}
        \includegraphics[width=1.0\linewidth]{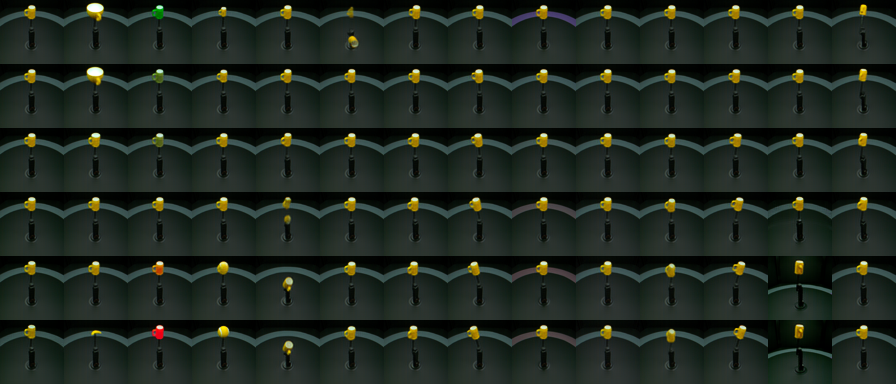}
    \end{subfigure}
    \hfill
    \begin{subfigure}{0.48\textwidth}
        \includegraphics[width=1.0\linewidth]{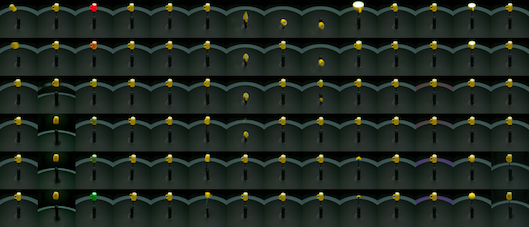}
    \end{subfigure}

    \begin{subfigure}{0.48\textwidth}
        \centering
        \includegraphics[width=1.0\linewidth]{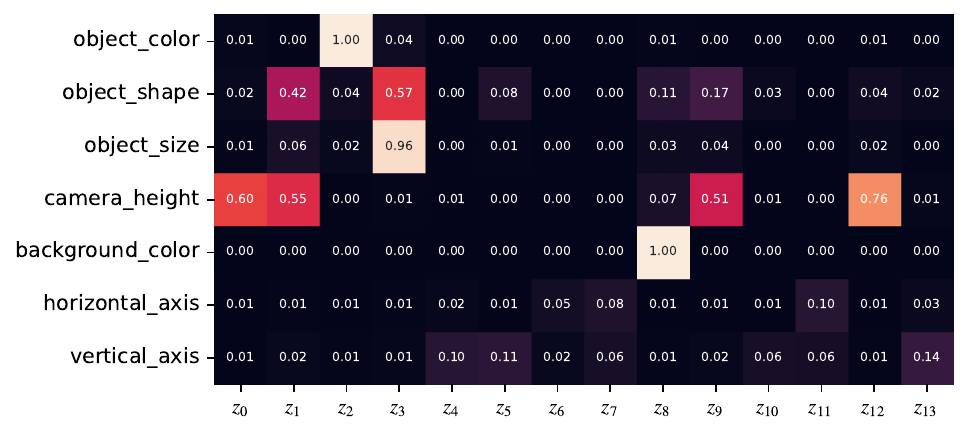}
        \caption{\ourmethod{} }
    \end{subfigure}
    \hfill
    \begin{subfigure}{0.48\textwidth}
        \centering
        \includegraphics[width=1.0\linewidth]{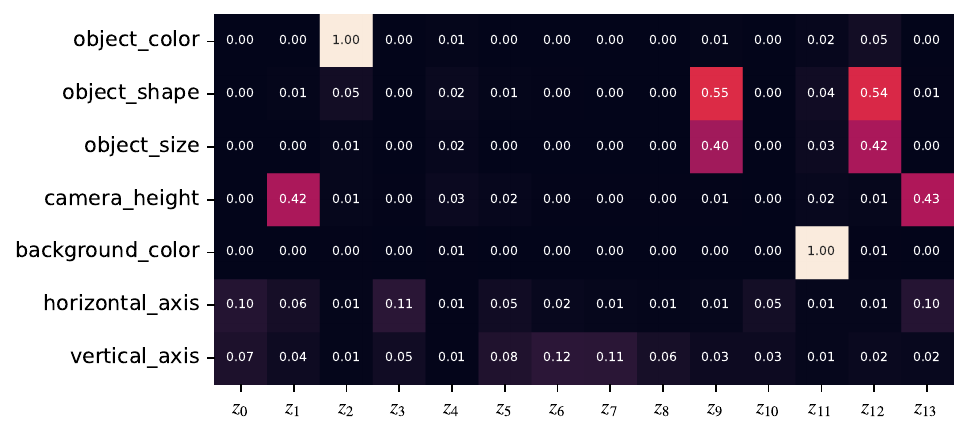}
        \caption{Naive \ourmethod{} }
    \end{subfigure}
    \caption{\ourmethod{} and naive \ourmethod{} decoded latent interventions and normalized mutual information heatmaps on MPI3D.}
\end{figure}

\vfill

\newpage

\begin{figure}[H]
    {\scriptsize
    \begin{subfigure}{0.48\textwidth}
        \begin{tabularx}{0.978\textwidth}{*{14}{>{\centering\arraybackslash}X}}
            \textcolor{red}{$z_0$} & $z_1$ & \textcolor{red}{$z_2$} & $z_3$ & $z_4$ & $z_5$ & $z_6$ & $z_7$ & $z_8$ & \textcolor{red}{$z_9$} & $z_{10}$ & $z_{11}$ & $z_{12}$ & \textcolor{red}{$z_{13}$} \\
        \end{tabularx}
    \end{subfigure}
    \hfill
    \begin{subfigure}{0.48\textwidth}
        \begin{tabularx}{0.978\textwidth}{*{14}{>{\centering\arraybackslash}X}}
            $z_0$ & $z_1$ & $z_2$ & $z_3$ & $z_4$ & $z_5$ & $z_6$ & $z_7$ & $z_8$ & $z_9$ & $z_{10}$ & $z_{11}$ & $z_{12}$ & $z_{13}$ \\
        \end{tabularx}
    \end{subfigure}
    }
    \begin{subfigure}{0.48\textwidth}
        \includegraphics[width=1.0\linewidth]{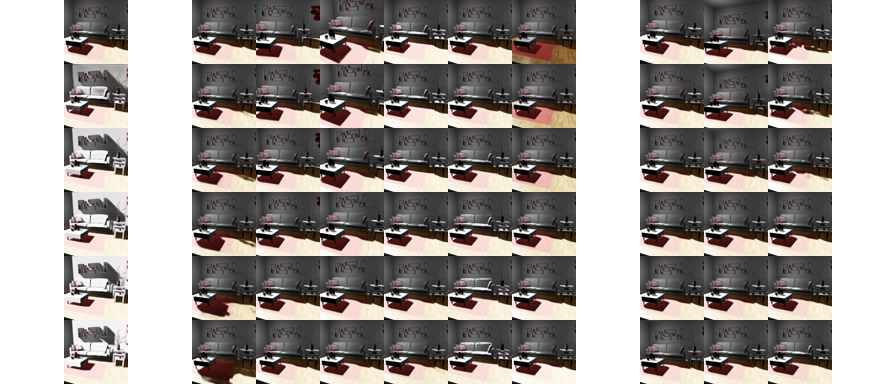}
    \end{subfigure}
    \hfill
    \begin{subfigure}{0.48\textwidth}
        \includegraphics[width=1.0\linewidth]{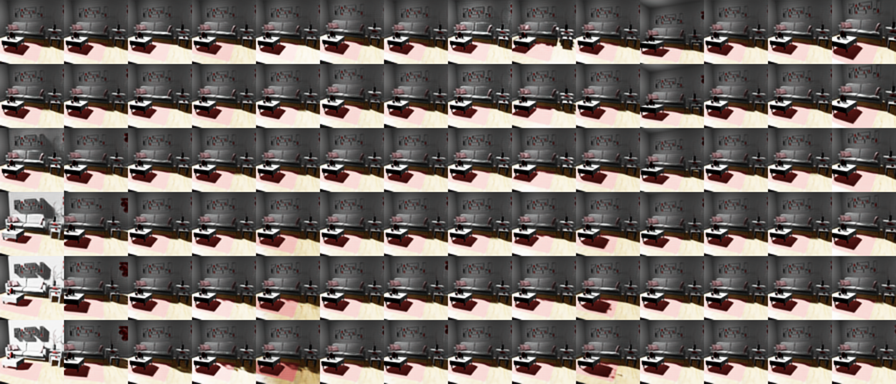}
    \end{subfigure}
    \begin{subfigure}{0.48\textwidth}
        \includegraphics[width=1.0\linewidth]{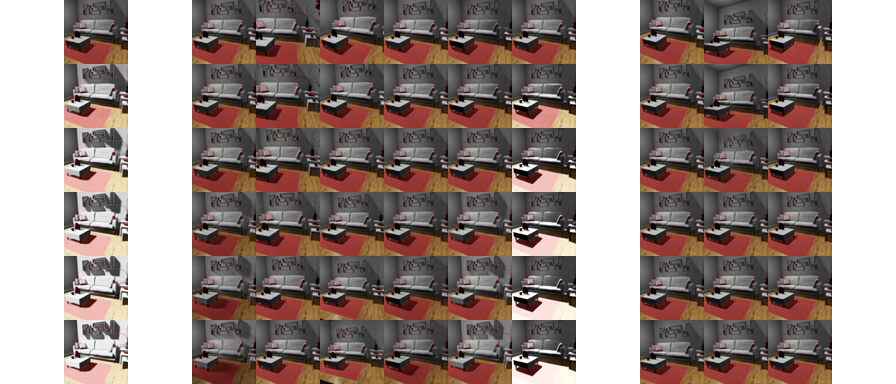}
    \end{subfigure}
    \hfill
    \begin{subfigure}{0.48\textwidth}
        \includegraphics[width=1.0\linewidth]{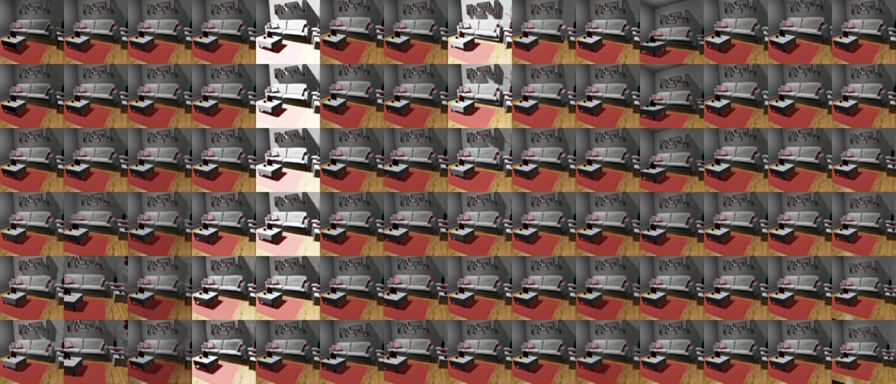}
    \end{subfigure}
    \begin{subfigure}{0.48\textwidth}
        \includegraphics[width=1.0\linewidth]{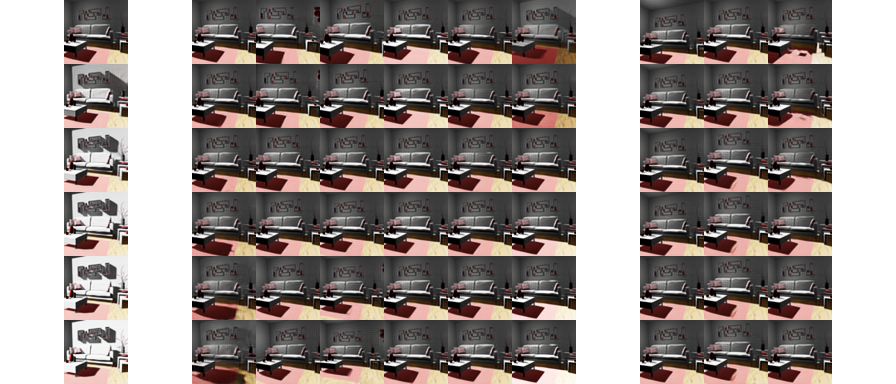}
    \end{subfigure}
    \hfill
    \begin{subfigure}{0.48\textwidth}
        \includegraphics[width=1.0\linewidth]{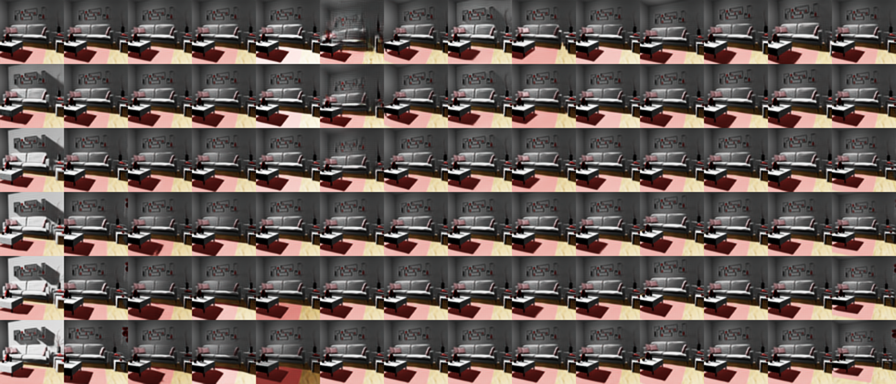}
    \end{subfigure}
    \begin{subfigure}{0.48\textwidth}
        \centering
        \includegraphics[width=1.0\linewidth]{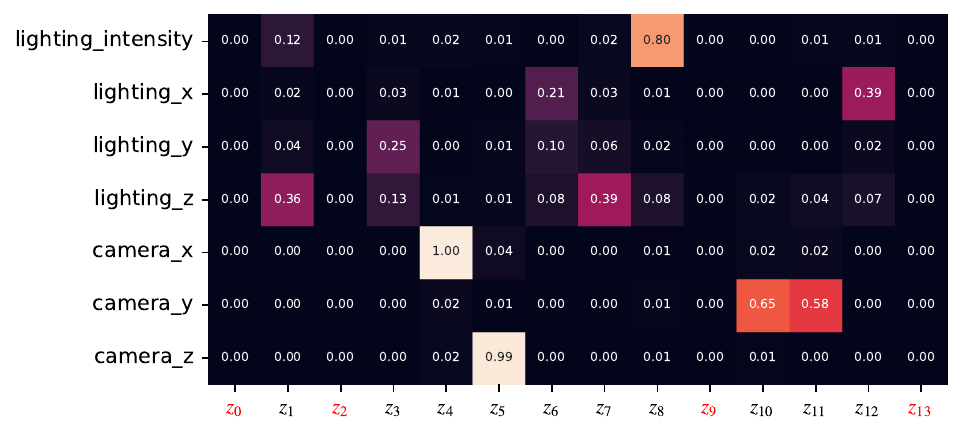}
        \caption{\ourmethod{}}
    \end{subfigure}
    \hfill
    \begin{subfigure}{0.48\textwidth}
        \centering
        \includegraphics[width=1.0\linewidth]{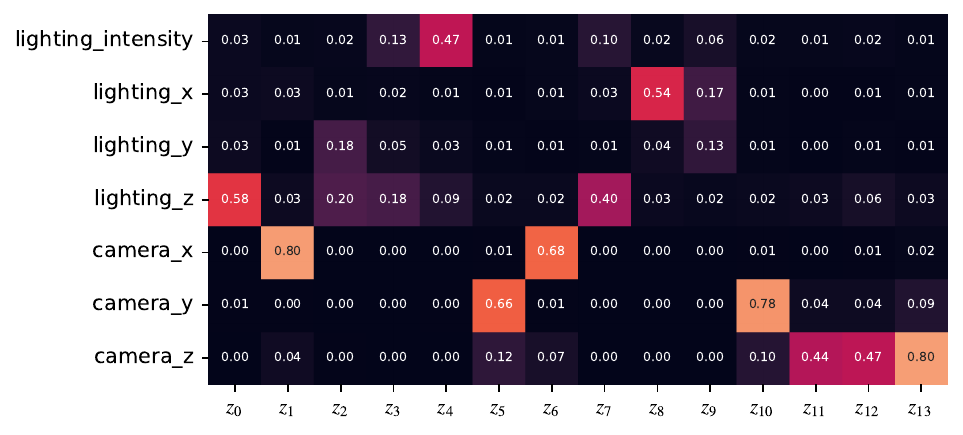}
        \caption{Naive \ourmethod{} }
    \end{subfigure}
    \caption{\ourmethod{} and naive \ourmethod{} decoded latent interventions and normalized mutual information heatmaps on Falcor3D.}
\end{figure}

\vfill

\newpage

\begin{figure}[H]
    {\scriptsize
    \begin{subfigure}{0.48\textwidth}
        \begin{tabularx}{0.97\textwidth}{*{18}{>{\centering\arraybackslash}X}}
            $z_0$ & \textcolor{red}{$z_1$} & $z_2$ & \textcolor{red}{$z_3$} & \textcolor{red}{$z_4$} & $z_5$ & $z_6$ & $z_7$ & $z_8$ & $z_9$ & $z_{10}$ & $z_{11}$ & $z_{12}$ & $z_{13}$ & $z_{14}$ & $z_{15}$ & $z_{16}$ & $z_{17}$ \\
        \end{tabularx}
    \end{subfigure}
    \hfill
    \begin{subfigure}{0.48\textwidth}
        \begin{tabularx}{0.97\textwidth}{*{18}{>{\centering\arraybackslash}X}}
            $z_0$ & $z_1$ & $z_2$ & $z_3$ & $z_4$ & $z_5$ & $z_6$ & $z_7$ & $z_8$ & $z_9$ & $z_{10}$ & $z_{11}$ & $z_{12}$ & $z_{13}$ & $z_{14}$ & $z_{15}$ & $z_{16}$ & $z_{17}$ \\
        \end{tabularx}
    \end{subfigure}
    }
    \begin{subfigure}{0.48\textwidth}
        \includegraphics[width=1.0\linewidth]{images/low_res/isaac_tripod_11.png}
    \end{subfigure}
    \hfill
    \begin{subfigure}{0.48\textwidth}
        \includegraphics[width=1.0\linewidth]{images/low_res/isaac_naive_step11_new.png}
    \end{subfigure}
    \begin{subfigure}{0.48\textwidth}
        \includegraphics[width=1.0\linewidth]{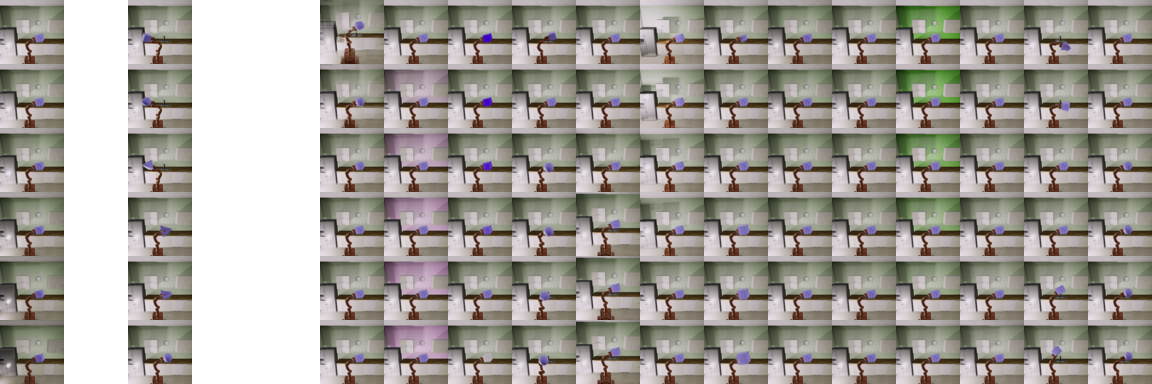}
    \end{subfigure}
    \hfill
    \begin{subfigure}{0.48\textwidth}
        \includegraphics[width=1.0\linewidth]{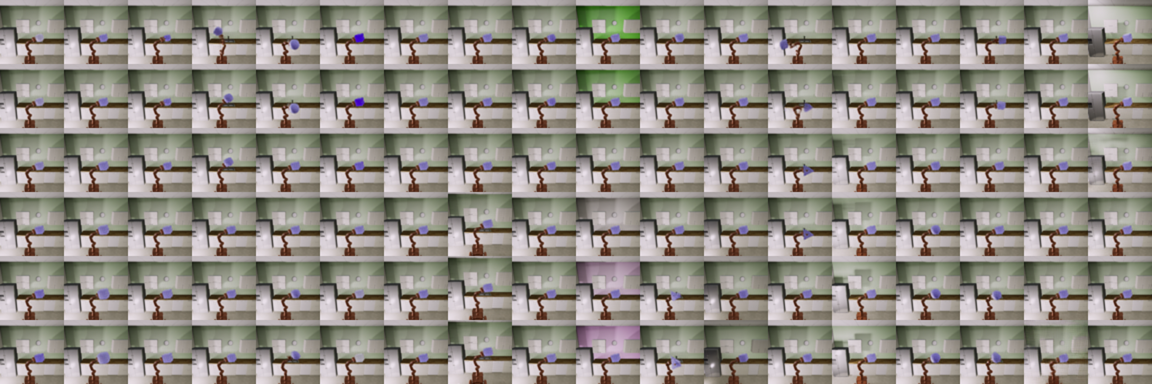}
    \end{subfigure}
    \begin{subfigure}{0.48\textwidth}
        \includegraphics[width=1.0\linewidth]{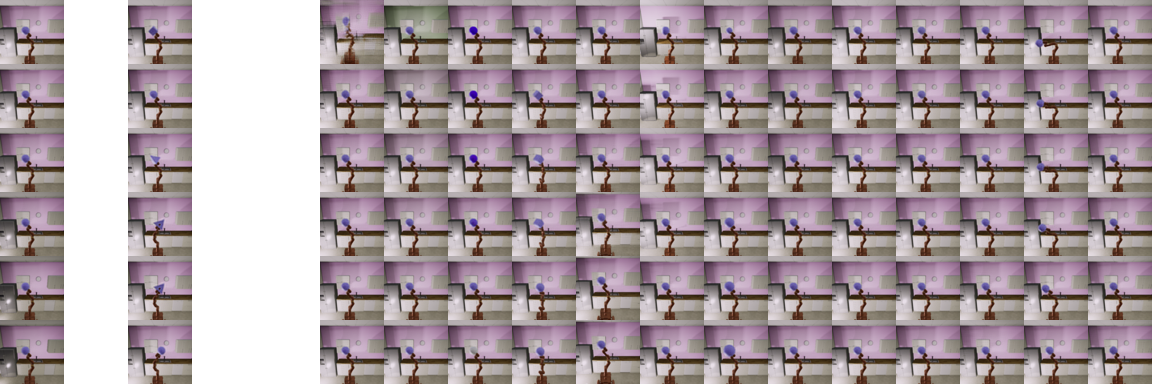}
    \end{subfigure}
    \hfill
    \begin{subfigure}{0.48\textwidth}
        \includegraphics[width=1.0\linewidth]{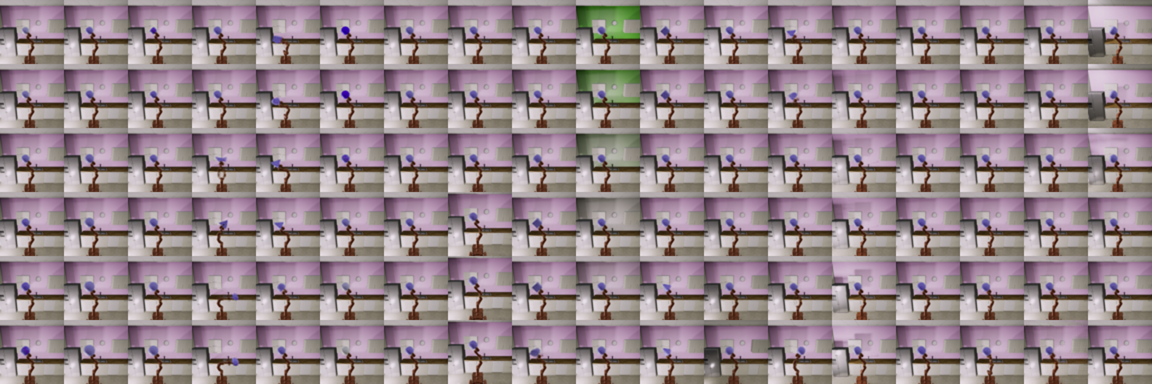}
    \end{subfigure}
    \begin{subfigure}{0.48\textwidth}
        \centering
        \includegraphics[width=1.0\linewidth]{images/low_res/isaac3d_tripod_corrected.pdf}
        \caption{\ourmethod{}}
    \end{subfigure}
    \hfill
    \begin{subfigure}{0.48\textwidth}
        \centering
        \includegraphics[width=1.0\linewidth]{images/isaac3d_naive.pdf}
        \caption{Naive \ourmethod{} }
    \end{subfigure}
    \caption{\ourmethod{} and naive \ourmethod{} decoded latent interventions and normalized mutual information heatmaps on Isaac3D.}
\end{figure}

\vfill

\newpage

\newpage
\section{Profiling Study}    \label{app:profiling_study}
We measure the average time (in seconds) each training iteration takes for various models. We observe that latent quantization and kernel-based latent multiinformation incur minimal overhead. However, adding normalized Hessian penalty increases the runtime by a factor of about $2.5$ due to the extra forward passes required to compute its regularization term. 
\begin{table}[h]
\centering
\begin{tabular}{lc}
\toprule
model & training iteration runtime (s) \\
\midrule
$\beta$-TCVAE & 0.040 \\
QLAE & 0.040 \\
QLAE $+$ KLM & 0.040 \\
Tripod & 0.106 \\
\bottomrule
\end{tabular}
\caption{Profiling study results.}
\label{tab:runtime}
\end{table}

\end{document}